\pdfoutput=1

\documentclass[11pt]{article}

\usepackage{naacl2021}

\usepackage{times}
\usepackage{latexsym}
\usepackage{colortbl}
\usepackage{amsmath}
\usepackage{graphicx}
\usepackage{verbatim}
\usepackage{multirow}
\usepackage{amssymb}

\usepackage[T1]{fontenc}
\usepackage[utf8]{inputenc}

\usepackage{microtype}

\title{The structure of online social networks modulates the rate of lexical change}

\author{Jian Zhu \\
  Department of Linguistics \\
  University of Michigan \\
  \texttt{lingjzhu@umich.edu} \\\And
  David Jurgens \\
  School of Information \\
  University of Michigan \\
  \texttt{jurgens@umich.edu} \\}

\begin{document}
\maketitle
\begin{abstract}
New words are regularly introduced to communities, yet not all of these words persist in a community's lexicon. Among the many factors contributing to lexical change, we focus on the understudied effect of social networks. We conduct a large-scale analysis of over 80k neologisms in 4420 online communities across a decade. Using Poisson regression and survival analysis, our study demonstrates that the community's network structure plays a significant role in lexical change. Apart from overall size, properties including dense connections, the lack of local clusters and more external contacts promote lexical innovation and retention. Unlike offline communities, these topic-based communities do not experience strong lexical levelling despite increased contact but accommodate more niche words. Our work provides support for the sociolinguistic hypothesis that lexical change is partially shaped by the structure of the underlying network but also uncovers findings specific to online communities. 

\end{abstract}

\section{Introduction}

Lexical change is a prevalent process, as new words are added, thrive, and decline in day-to-day usage. While there is a certain randomness at play in word creation and adoption \cite{newberry2017detecting}, there are also psychological, social, linguistic and evolutionary factors that systematically affect lexical change \cite{labov2007transmission,christiansen2003language,lupyan2010language}. 

In sociolinguistics, one structural factor that has long been recognized as influencing lexical changes is the language community's social network. For example, drawing on pioneering works on social networks  \cite{granovetter1977strength,granovetter1983strength}, the {\it weak tie model of change} holds that the structural properties of social networks can account for the general tendency of some language communities to be more resistant to linguistic change than others \cite{milroy1985linguistic,milroy1992social,milroy2013social}. A classic finding is that loose-knit networks with mostly weak ties are more conducive to information diffusion, thereby facilitating innovation and change, while close-knit networks with strong bonds impose norm-enforcing pressure on language usage, strengthening the localized linguistic norms \cite{milroy1985linguistic}. 

One compelling observation in favor of this argument concerns the comparison between two Germanic languages, Icelandic and English.  Icelandic has changed little since the late thirteenth century, which could be due to the norm-enforcing pressure inherent in the strong kinship and friendship ties. In contrast, in Early Modern London English, the loosening of network ties, accompanied by the rise of the mobile merchant class, was argued to be responsible for some radical change in the language \cite{milroy1985linguistic}.

\begin{figure*}[tbh]
    \centering
    \includegraphics[width=0.48\textwidth]{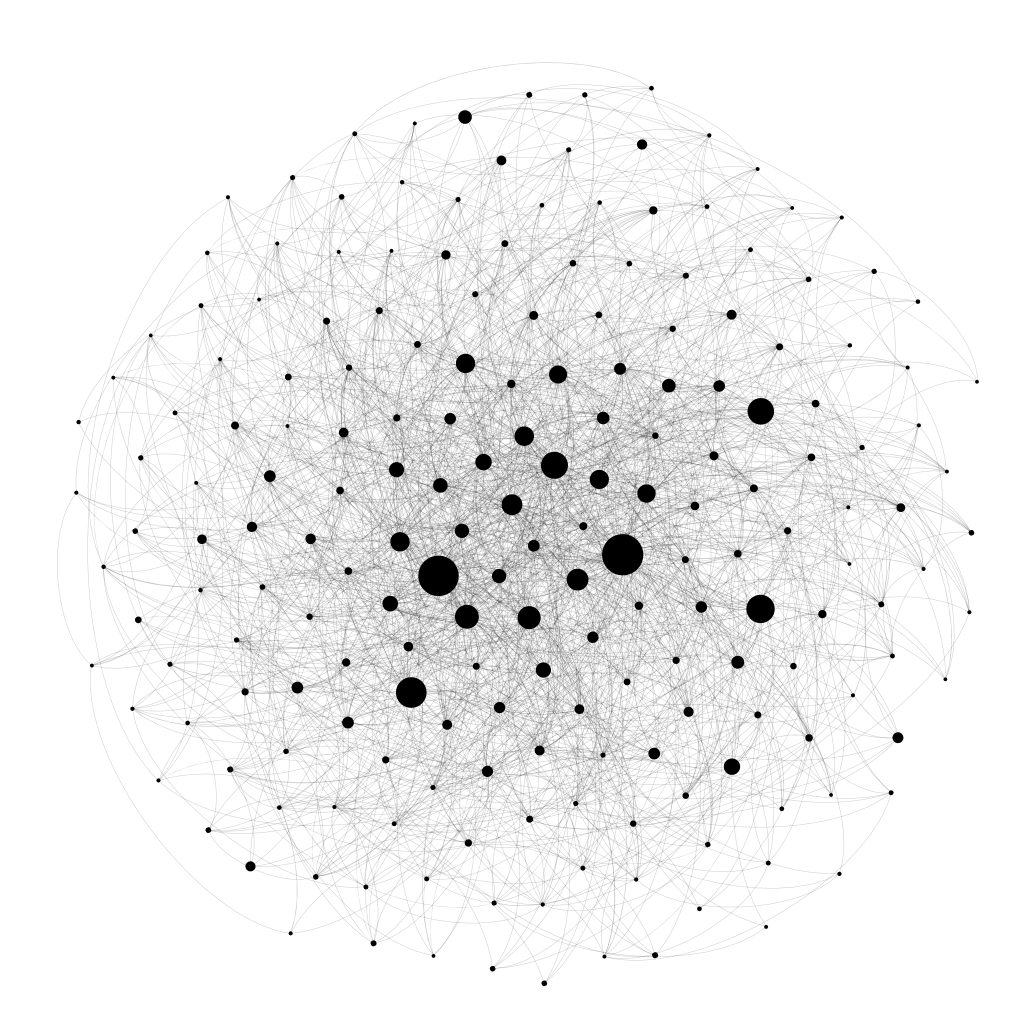}
    \includegraphics[width=0.48\textwidth]{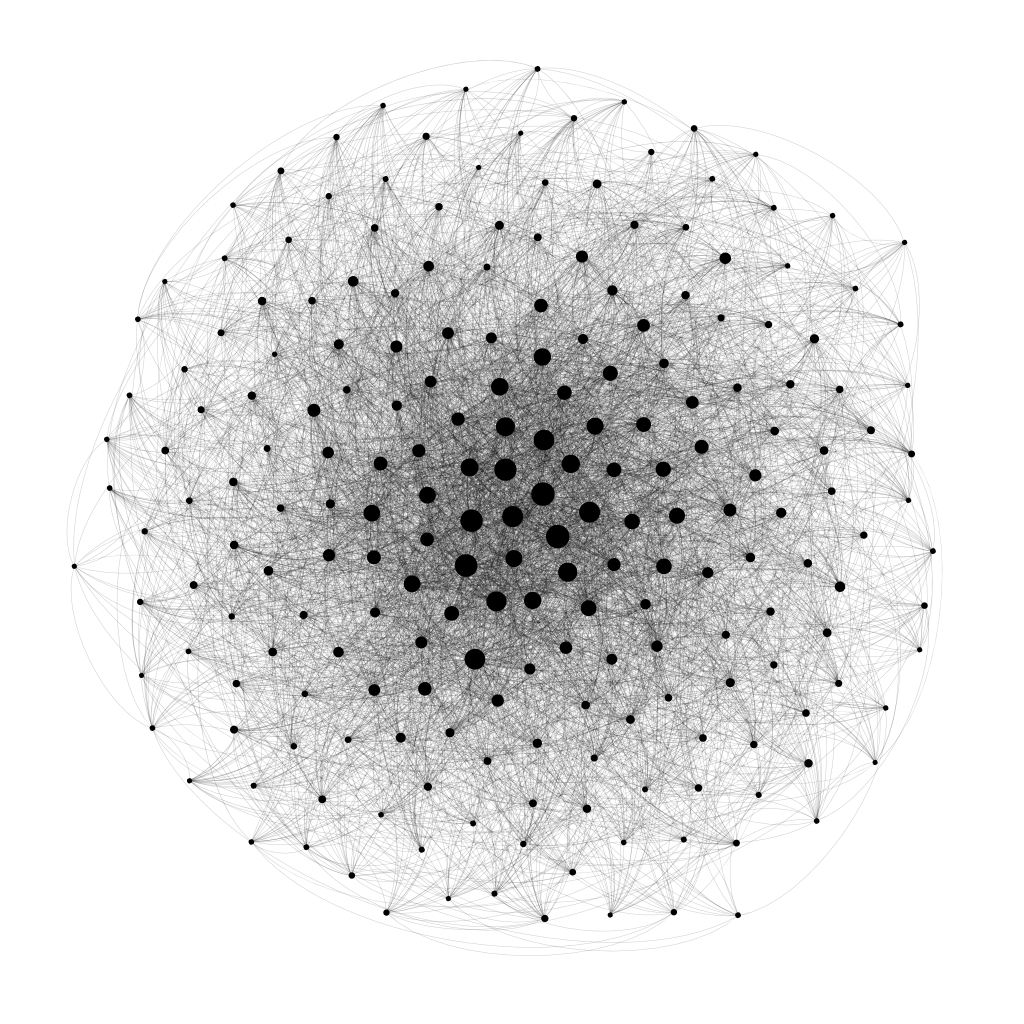}
    \caption{Applying the hypothesis of \citet{milroy1985linguistic} to these two gaming subreddits of similar size suggests that the network with lower density (left; \texttt{r/masseffect}) will be more innovative than the more closely-connected community shown right (\texttt{r/F13thegame}). However, after controlling for size, the one with higher average degree (more inner-connections) (right: \texttt{r/F13thegame}) tends to develop more lexical innovations. 
    %
    }
    \label{fig:networks}
\end{figure*}

This study extends network-based sociolinguistic research to online communities, which remain understudied despite their expansion in past decades. While we draw an analogy between offline and online communities, our focus is on communities of practice \cite{eckert1992think,holmes1999community,schwen2003community}, or ``an aggregate of people who come together around mutual engagement in an endeavor'' \cite{eckert1992think}, rather than offline speech communities. We examine how network structures affect lexical {\bf innovation}, {\bf retention} and {\bf levelling} in online communities. Specifically, we ask 1) how network structure contributes to the introduction of new words to online communities (innovation), 2) how structural properties affect the survival of these newly introduced words (retention) and 3) whether the increased inter-connectedness causes online communities to adopt a similar set of new words (levelling).

This work offers the following contributions. First, using a massive longitudinal dataset of 4420 communities, we precisely quantify the structural mechanisms that drive these lexical processes. Our work adds to network studies in sociolinguistics focusing on in-person observations of local communities \cite{conde201218,sharma2020language} and shows that conclusions drawn from offline communities are insufficient to account for behavior seen in online social networks (Figure~\ref{fig:networks}). We find that larger size, denser connections, lack of local clustering and greater external contacts promote lexical innovation and retention in online communities, while density, as discussed most in offline studies, could be an emergent byproduct of network size. These topic-based communities also do not experience strong levelling due to increased contact. 
Second, emerging studies in online communities \cite{danescu2013no,stewart-eisenstein-2018-making,del-tredici-fernandez-2018-road} focus exclusively on lexical change at the individual or word level. Few investigate how global network properties affect lexical change at the community level. Finally, sampling offline networks presents practical difficulties, we extract complete networks for thousands of online communities, providing a large-scale dataset to explore the structural factors of lexical change. Our code is available at \url{https://github.com/lingjzhu/reddit_network} and replication details are available in Appendix~\ref{replic}.

\section{Lexical Change}

\paragraph{Lexical change and social networks}
Since the landmark study of sound change in the Belfast community by \citet{milroy1985linguistic}, the impact of network structures on language change has been a key consideration in sociolinguistics.  \citet{milroy1985linguistic} found that speakers in loose-knit networks tend to experience more linguistic change than those in close-knit networks. 
Most early social network studies focus predominantly on speakers in local, less mobile communities where ties between people tend to be strong \cite{nevalainen2000mobility,conde201218,sharma2020language}. Except for a few recent simulation studies \cite{reali2018simpler}, researchers have rarely explored how the global properties of social networks systematically affect lexical change, although the weak tie model does predict an influence of social network at the macro-level. In addition, while there are lexicographic studies attempting to enumerate factors that affect the acceptance of neologisms \cite{metcalf2004predicting,barnhart2007calculus}, network structures are rarely taken into consideration. A key limitation of previous works has been access to a large longitudinal dataset of communities with different network properties as well as a precise estimate of the network structure of larger communities, which are limitations this study overcomes.

\paragraph{Lexical change in online communities}
The rise of social media and the proliferation of Internet speech has drawn increasing attention to lexical change in online communities, including Twitter \cite{eisenstein2014diffusion, goel2016social},  Reddit \cite{altmann2011niche,stewart-eisenstein-2018-making,del-tredici-fernandez-2018-road} and review sites \cite{danescu2013no}. It has been shown that the usage of certain words is associated with community loyalty and norms \cite{zhang2017community,bhandari-armstrong-2019-tkol} and indicative of user behaviors \cite{danescu2013no,noble-fernandez-2015-centre,chang2019trajectories,klein2019pathways}.  Specifically for lexical change over time, \citet{stewart-eisenstein-2018-making} investigate the survival of lexical items in Reddit, and conclude that a word's appearance in more diverse linguistic contexts is the strongest predictor of its survival while social dissemination is a comparatively weaker predictor. \citet{del-tredici-fernandez-2018-road}  examined the use of neologisms in 20 subreddit communities. Their finding that weak-tie users tend to innovate whereas strong-tie users tend to propagate is consistent with \textit{the weak tie theory of language change}. Other studies along this line tend to focus on the role of individual users \cite{paolillo1999virtual,paradowski2012diffusion}. The study closest to our current study is that by \citet{kershaw2016towards}, which investigates word innovations in Reddit and Twitter by looking at grammatical and topical factors. Yet \citet{kershaw2016towards} only used network information to partition the dataset without exploring the role of these structural attributes in depth. Less is known about how network structures are systematically related to community-level lexical change in online communities, which we address here.

\section{The Reddit Network Corpus}

To analyze lexical innovation in a network setting across long time scales, we use comments made to Reddit, one of the most popular social media sites. There, 330M users are active in about 1M distinct topic-based sub-communities (subreddits). 
Here we define each subreddit as a community of practice \citep{schwen2003community}, as each subreddit is relatively independent with various norms formed through interactions. The subreddit communities span across a wide range of social network structures \cite{hamilton2017loyalty} and linguistic use patterns \cite{zhang2017community}, making them ideal for studying the propagation of sociolinguistic variations in online communities.  Detailed statistics are given in Appendix~\ref{app: corpus}.


\paragraph{Data} To strike a balance between acquiring active subreddits and preserving the diversity of these communities, we initially select the top 4.5K subreddits based on their overall size from their inception to October 2018 via the \texttt{Convokit} package \cite{chang2020convokit}.
Let $\mathcal{C}_{Reddit}=\{C_1,C_2, \dots, C_n\}$ be the set of subreddit communities included in the corpus. A subreddit community $C_n$ is further discretized into multiple monthly subreddit communities $c_n(t)$ based on its actual life span in the monthly time step $t$, such that $C_n = \{c_n(1), c_n(2), \dots, c_n(t_{max})\}$. For each $c_n(t)$, we extracted all individual comments except those marked as \texttt{[deleted]} and performed tokenization via SpaCy. 
During text cleaning, we removed numbers, emojis, urls, punctuations and stop words, and set a cutoff frequency of 10 over the entire dataset to exclude infrequent typos or misspellings. 
Only those monthly subreddits $c_n(t)$ with more than 500 words or 50 users after preprocessing are retained. Some communities known for their content in foreign languages are also removed. After preprocessing, 4420 subreddits were left in our analysis.   

\paragraph{Community networks}  For a community $c$ from month $t=1,2,\dots,t_{max}$, its temporal network can be represented as a discrete-time sequence of network snapshots $\mathcal{G}_c = \{G_c(1), G_c(2), \dots, G_c(t_{max})\}$. Each snapshot network at time $t$,  $G_c(t)=\{V_c(t), E_c(t)\}$ consists of a set of user nodes $V_c(t)$ and a set of edges $E_c(t)$ characterizing direct interactions between users. $G_c(t)$ is initiated as an undirected and unweighted graph under the assumption that these commenting communications are mutual and bi-directional. 

A user $u_i$ is represented as a node if this user has posted at least one comment at month $t$. An edge $e_{ij}$ exists between user $u_i$ and user $u_j$ if these two users have interacted in close proximity in a common discussion thread, that is, separated by at most two comments \cite{hamilton2017loyalty,del-tredici-fernandez-2018-road}.  Since online communications are asynchronous, a discussion thread created at time $t$ may still have active comments from users at time $t+1$ or later. For such threads, we only included interactions at time $t$ in $G_c(t)$ and grouped later interactions into the future time steps at which these interactions happened. Users marked as \texttt{[deleted]} or \texttt{AutoModerater} were all removed. 
After filtering, a total of 289.8k community networks have been extracted for all 4420 communities.

\paragraph{Inter-community networks}
We also identify the network dynamics between communities. 
We created temporal network $\mathcal{G}_{IC}$ to characterize the connections between communities at consecutive months  $t=1,2,\dots,t_{max}$, $\mathcal{G}_{IC}=\{G_{IC}(1), G_{IC}(2), \dots, G_{IC}(t_{max})\}$, in which $G_{IC}(t)=\{V_{IC}(t), E_{IC}(t)\}$.  $V_{IC}(t)$ contains the set of nodes whereas $E_{IC}(t)$ is the set of edges between communities. A community is represented as a node $u_i$ in $G_{IC}(t)$, except for communities that do not exist or are no longer active at time $t$. Two communities are determined to be connected if they share active users, that is, users who had posted at least 2 comments in both communities during that month. Each network snapshot is initiated as a weighted and undirected network with the edge weights set to the numbers of shared users, as an approximation of connection strength. Finally, 152 inter-community networks have been constructed since the inception of Reddit in 2005 until October 2018.

\paragraph{Internet neologisms}
Neologisms are newly emerging language norms that fall along a continuum from the common words known to the overwhelming majority of users to nonce words that are mostly meaningless and rarely adopted. We only focus on Internet neologisms, e.g. \textit{lol, lmao, idk}, as community slangs in Reddit communities. Such neologisms are abundant in the ever-evolving online communications as people use them for convenience or to signify in-group identity.
The non-standard, idiosyncratic spelling patterns of Internet neologisms also make them easier to track than nuanced meaning shifts. 

We obtained the Internet slangs from two online dictionary sources, \texttt{NoSlang.com} and \texttt{Urban Dictionary}. The neologisms in \texttt{NoSlang.com} have been used in a previous study \cite{del-tredici-fernandez-2018-road}.  
After filtering some lexical entries, we ended up with approximately 80K Internet neologisms for subsequent analysis. We set the minimum frequency threshold of neologisms to 10 over the entire dataset; this low setting ensures that the analysis is not biased by selectively looking only at surviving words, which may obscure the lexical change process. Details can be found in Appendix~\ref{app: corpus}. 

Many of these neologisms were not first coined in Reddit but were coined elsewhere and introduced into subreddits subsequently by users. Since it was neither feasible nor possible to trace the exact origins of these words, we instead focused on how words were introduced and adopted. This approach is also consistent with previous studies of lexical change \cite{altmann2011niche,grieve2017analyzing,del-tredici-fernandez-2018-road}. 

\begin{table}[tb]
\small
\centering
\begin{tabular}{ll}
\textbf{Frequency}  & \textbf{Neologisms} \\ \hline
\shortstack{Most frequent}
            & \shortstack[l]{{ lol, /r, kinda, bitcoin, idk, lmao, tbh} \\
{ tl;dr, alot, /s, omg, lvl, hahaha, iirc}} \\ \hline
\shortstack{Least frequent} & \shortstack[l]{{ thugmonster, blein, sotk, f'tang} \\
                                { yobbish, ferranti, sonse, yampy}}     \\

\end{tabular}
\caption{Examples of neologisms.}
\label{table:exp_neo}
\end{table}


\section{Network statistics}\label{net_stats}

Communities in Reddit can be defined in terms of how their members relate within the community (intra) and how the community relates to other communities (inter) through multi-community memberships by its users \citep{tan2015all}. We formalize both as potential influences. As network attributes may be affected by the hyperparameters for network construction, we additionally validate this approach in Appendix~\ref{app: network}.


\paragraph{Intra-community features} 
We  take the following network measurements for each $G_c(t)$ to characterize the global properties of community networks: density, average local clustering coefficient, transitivity, average degree, maximum degree, degree assortativity, fraction of the largest connected components and fraction of singletons. These network measures can characterize the size, fragmentation and connectedness of Reddit networks \cite{hamilton2017loyalty,cunha2019all}.

Parameters like average local clustering coefficient, transitivity, and assortativity are highly influenced by the underlying degree distribution \cite{hamilton2017loyalty}. We adjusted these parameters by computing their relative differences with respect to the mean values of five random baseline networks, which were generated by randomly rewiring the original network for 10 $\times$ edge count iterations and preserving the original degree sequence. These features are referred to as adjusted local clustering coefficient, adjusted transitivity, and adjusted assortativity in the following text. 


\paragraph{Inter-community features}
In addition to the intra-community network features, it is also necessary to measure a community's external connections to other communities. User mobility and external influence have been found to play a role in the process of lexical change \cite{conde201218}. 
For each between-community network snapshot $G_{IC}(t)$ at time $t$, we focus on the properties of individual nodes (communities). We computed the degree centrality, closeness centrality, eigenvector centrality, betweenness centrality and PageRank centrality for each community node. These centrality measures quantify the connectedness of a community to other communities, which can be used as an indicator of their degree of external contact and user mobility.

\begin{figure*}[tbh]
    \centering
    \includegraphics[width=\textwidth]{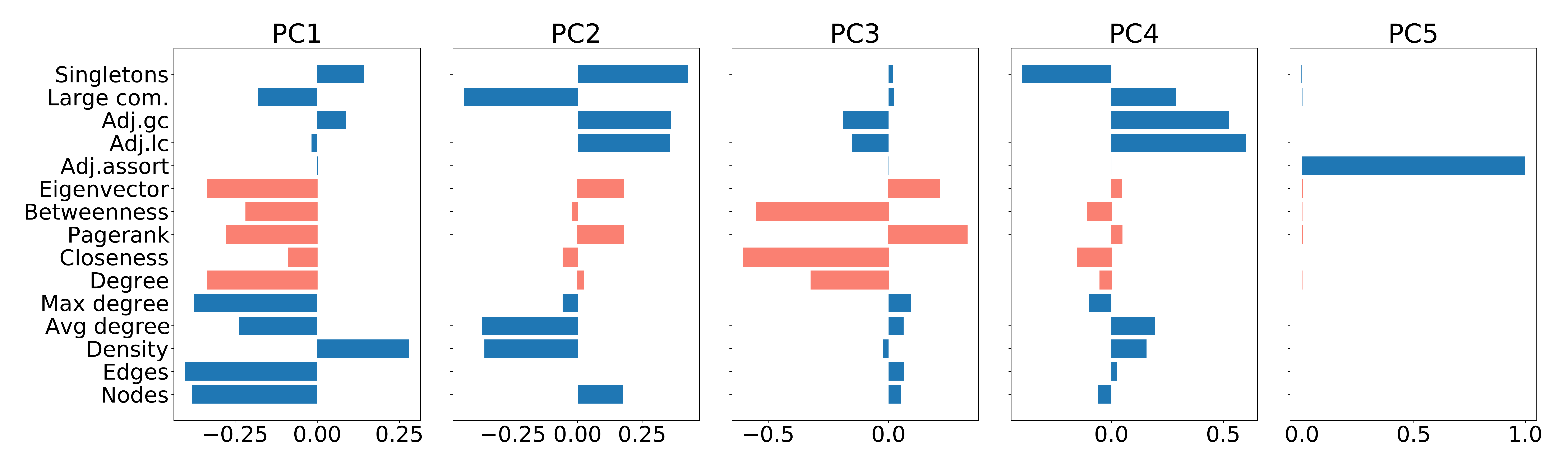}
    \caption{The decomposition of the PCs used in predicting innovations. Inter-community features are highlighted in orange bars. Adj.lc, Adj.gc and Adj.assort are local clustering coefficient, global clustering coefficient and assortativity adjusted with respect to a random network. PC1 represents the overall size, PC2 the density of intra-community connections and PC3 the inter-community connections.}
    \label{fig:pcs_innovation}
\end{figure*}

\section{Lexical innovations}

In what types of communities are neologisms likely to be introduced? 
Here, we investigate the extent to which the number of innovations introduced per month can be predicted with only the structural properties of community networks. 

\paragraph{Experiment setup} Given a set of communities $C=\{c_1, c_2, \dots, c_n\}$ spanning time steps $T=\{1,2,\dots,t_{max}\}$, we aim to predict the count of monthly lexical innovations for each community $Y=\{y_{1}^{c_1}, y_{2}^{c_1}, \dots, y_{t_{max}}^{c_n}\} $ from the corresponding network attributes ${\bf X}=\{{\bf x}_1^{c_1},{\bf x}_2^{c_1},\dots,{\bf x}_{t_{max}}^{c_n}\}$. The predicted variable $y_t^{c_n}$ is computed by counting only innovations first introduced into community $c_n$ at month $t$. Any subsequent usage of the same innovations after their first introduction is not counted as innovations in community $c_n$. The feature vector ${\bf x}_t^{c_n}$ is the structural features of the network at time $t$ for $c_n$. After removing about 0.03\% invalid data points and outliers, we ended up with 289.1k samples for the task.

\paragraph{Implementation} We used both intra-community and inter-community features for innovation prediction. 
However, in empirical networks, certain structural features tend to be correlated. For example, network size and density are usually strongly correlated on a log-log scale in online social networks \cite{backstrom2012four}, which is also apparent in our dataset (Spearman $\rho$=-0.87).  Such correlations may confound the interpretation of the feature contributions (see Appendix~\ref{app:corr}). To generate orthogonal features, we first standardized all 15 network features and then used principal component analysis (PCA) with whitening to decompose them into principal components (PCs). Standardization was necessary as it could prevent a few variables with a large range of variance from dominating the PCs. We found that the first five PCs accounted for 87\% of total variance and 10 PCs explained 99\% of the total variance.     

Since counts of innovations are non-negative integers, Poisson regression and Histogram-based Gradient Boosted Trees (HGBT) with Poisson loss were used to predict the number of innovations with PCs. The model parameters were selected through ten-fold cross-validation. The data were randomly partitioned into training and test sets with a ratio of 90\%/10\%. We report the mean absolute error (MAE) and the mean Poisson deviance (MPD) averaged across 20 runs with different random partitions of data. Both metrics should be minimized by the models. Replication details are in Appendix~\ref{app:innovate}.

\begin{table}[tbh]
\small
\centering
\begin{tabular}{lrr}
\textbf{Model} & \textbf{MAE} & \textbf{MPD} \\ \hline
Baseline (mean) & 19.37 & 30.16 \\
Poisson reg. (PCs=5) & 11.79 & 12.29 \\
Poisson reg. (PCs=10) & 11.14 & 11.03 \\
Poisson reg. (raw feat.) & 11.72 & 12.21 \\ 
HGBT (PCs=5) & 10.57 & 9.63 \\
HGBT (PCs=10) & 9.65 & 8.19 \\
HGBT (raw feat.) & {\bf 9.24} & {\bf 7.49} \\
\end{tabular}
\caption{Results of lexical innovation prediction.}
\label{table:inno_pred}
\end{table}

\paragraph{Results} As summarized in Table~\ref{table:inno_pred}, all models outperformed the mean baseline by a significant margin, suggesting that the internal network structures and the external connections to other communities are systematically correlated to the count of lexical innovations per month. The three largest coefficients of the Poisson model with 5 PCs correspond to the first three PCs (see Figure~\ref{fig:pcs_innovation}) \footnote{Note that the coefficient sign for a PC must be interpreted with respect to to its loading on structural components.}. PC1 represents the overall size of the network, such that the Poisson model predicts that networks having larger overall size tend to 
have more innovations (Coefficient: -0.87). PC2 indicates the fragmentation and the local clusteredness of the network, and contributes negatively to lexical innovation (Coefficient.: -0.20). In other words, fragmented networks with local clusters tend to have fewer innovations as this structure inhibits the spread of information. PC3 is generally related to inter-community connections with positive correlation to innovation (Coefficient.: 0.19). Yet what matters is not the number of communities connected (degree centrality) but the quality of those connections (Pagerank centrality). High Ragerank centrality suggests that the network might be connected to many influential communities, as these connections are weighted higher in the Pagerank algorithm \cite{page1999pagerank}.

While structural properties can account for many regularities in the creation of lexical innovations, there are also surges of innovations that cannot be explained by structural factors alone. Inspection of the data suggests that the surges of innovations at the tail of empirical distributions are often related to some factors beyond network structures, including topical variations or external events, such as community migration or new game releases for some game communities.

\section{Survival Analysis}
Not all lexical innovations survive through time, with only a few neologisms eventually becoming widely adopted by community members.  Here, we test the structural factors that systematically affect the survival of words in online communities. 

\paragraph{Model specification} 
Survival analysis models the elapsed time before a future event happens \cite{kleinbaum2010survival}, which has been used to predict word survival \cite{stewart-eisenstein-2018-making}. Compared to the traditional Cox model, deep survival analysis approximates the risk (hazard) with neural networks, thereby achieving improved performance. We estimated word survival with the Logistic Hazard model (LH) proposed by \citet{kvamme2019continuous}. Given samples $\{{\bf x}_1, {\bf x}_2, \dots, {\bf x}_n\}$ and time steps $\{1, 2, \dots, T\}$, the LH method estimates $h(t|{\bf x})$, the hazard function of the death event with respect to time $t$, with a deep neural network. The hazard function can be interpreted as the word's ``danger of dying" at $t$. 

After the model is trained, the survival function $S(t|{\bf x}_i)$ for sample ${\bf x}_i$ can be computed as
\begin{equation}
    S(t|{\bf x}_i) = \prod_{t=1}^T[1-h(t|{\bf x}_i)]
\end{equation}
$S(t|{\bf x}_i)$ can be interpreted as the chance of survival at time $t$ for sample ${\bf x}_i$, that is, the survival probability of a word given the corresponding network features at time $t$. The detailed derivation and experiment settings are given in Appendix~\ref{app: survival}. 


\paragraph{Data coding} We consider only communities that have existed longer than six months and words that survived more than three months.
The subreddit duration restriction avoids right-censoring of the data from new communities forming and quickly dying (a common event), which would skew estimates of word survival.
A word's survival time is defined as the total number of months a word persists in a community, excluding the intervening month in which the word is not used.
The last time step $t$ at which the word shows up is considered the ``death" event. However, if this last time step is also the last three recorded months, this word is considered right-censored such that a death event has not happened. This three-month buffer period is added to avoid false negatives. 
The network features for predictions were derived from averaging all the monthly features for the months that a particular word has existed.  After preprocessing, we ended up with 1.47M samples with 69,683 distinct words. All features were then transformed into 10 orthogonal principal components using PCA with whitening. The first 5 PCs accounted for 90\% of the total variance  whereas all 10 PCs explained 99\% of the variance. 


\paragraph{Implementation} Models of deep survival analysis were implemented via the package \texttt{pycox} \cite{JMLR:v20:18-424}. 
We trained a three-layered LH model with 256 hidden dimensions to model the word survival. We used the Adam optimizer with a learning rate of 0.001 and a batch size of 2048 samples. The data were randomly partitioned into  80\%, 10\% and 10\% portions as training, development and test sets, respectively, with no overlap between sets in terms of subreddits. Each model was run for 3 epochs and was run 10 times with different data partitioning. The performance metrics were averaged. We also ran baseline Cox models under the same conditions for comparison.

The performance is evaluated with {\it time-dependent concordance} \cite{antolini2005time} and {\it Integrated Brier Score} (IBS) \cite{JMLR:v20:18-424}.  Concordance measures the model's capacity to provide a reliable ranking of individual risk scores. A good concordance score should be above the 0.5 random baseline and close to 1. The IBS is the average squared distances between the observed survival events and the predicted survival probability and should be minimized by the model. 


\begin{table}[tb]
\small
\centering
\begin{tabular}{lrr}
\textbf{Model} & \textbf{Concordance} & \textbf{IBS} \\ \hline
Random baseline & 0.50 & 0.25 \\
Cox Model (PCs=5) & 0.600 & 0.297 \\ 
Cox Model (PCs=10) & 0.662 & 0.289 \\ 
Cox Model (raw) & 0.665 & 0.209 \\ 
LH (PCs=5) & 0.584 & 0.245 \\
LH (PCs=10) & 0.691 & 0.192 \\
LH (raw) &  {\bf 0.718} &  {\bf 0.152} \\
\end{tabular}
\caption{Survival analysis results. All models outperform the concordance baseline.}
\label{table: survival}
\end{table}

\begin{figure*}[tb]
    \centering
    \includegraphics[width=\textwidth]{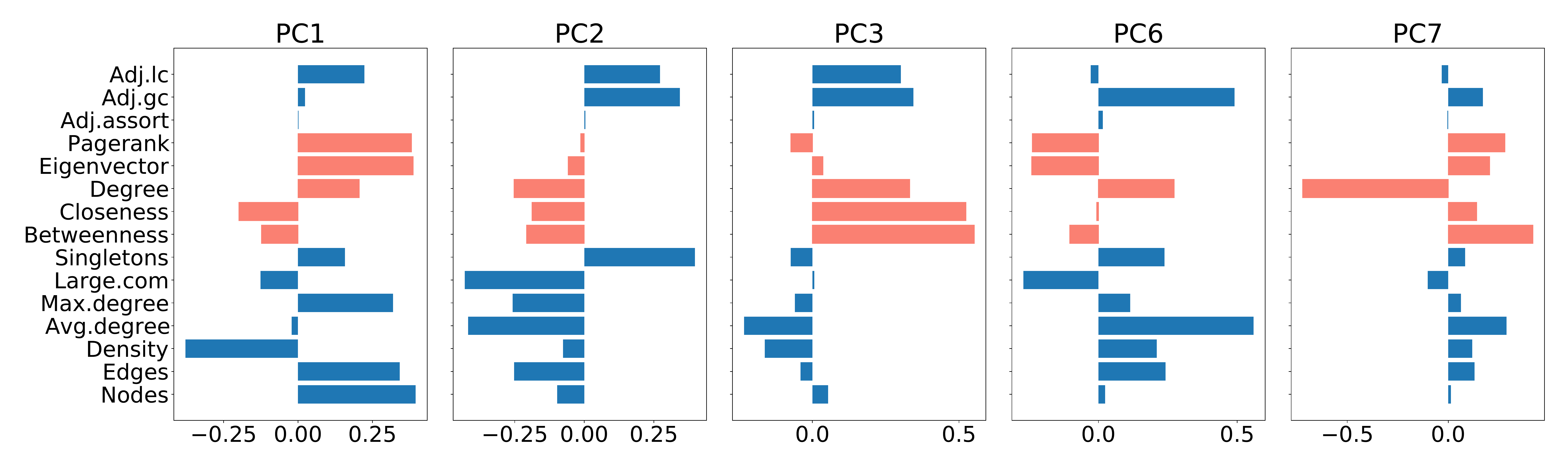}
    \caption{The five highest weighted PCs used by the survival model. Inter-community features are indicated by orange bars. Adj.lc, Adj.gc and Adj.assort are local clustering coefficient, global clustering coefficient and assortativity adjusted to a random network. PC1 represents the overall size, PC2 the within-community connections and PC3 the inter-community connections. PC6 and PC7 are specific combinations of intra- and inter-community connections.}
    \label{fig:pcs_survival}
\end{figure*}

\paragraph{Results} 

\begin{figure}[tb]
    \centering
    \includegraphics[width=\columnwidth,height=4cm]{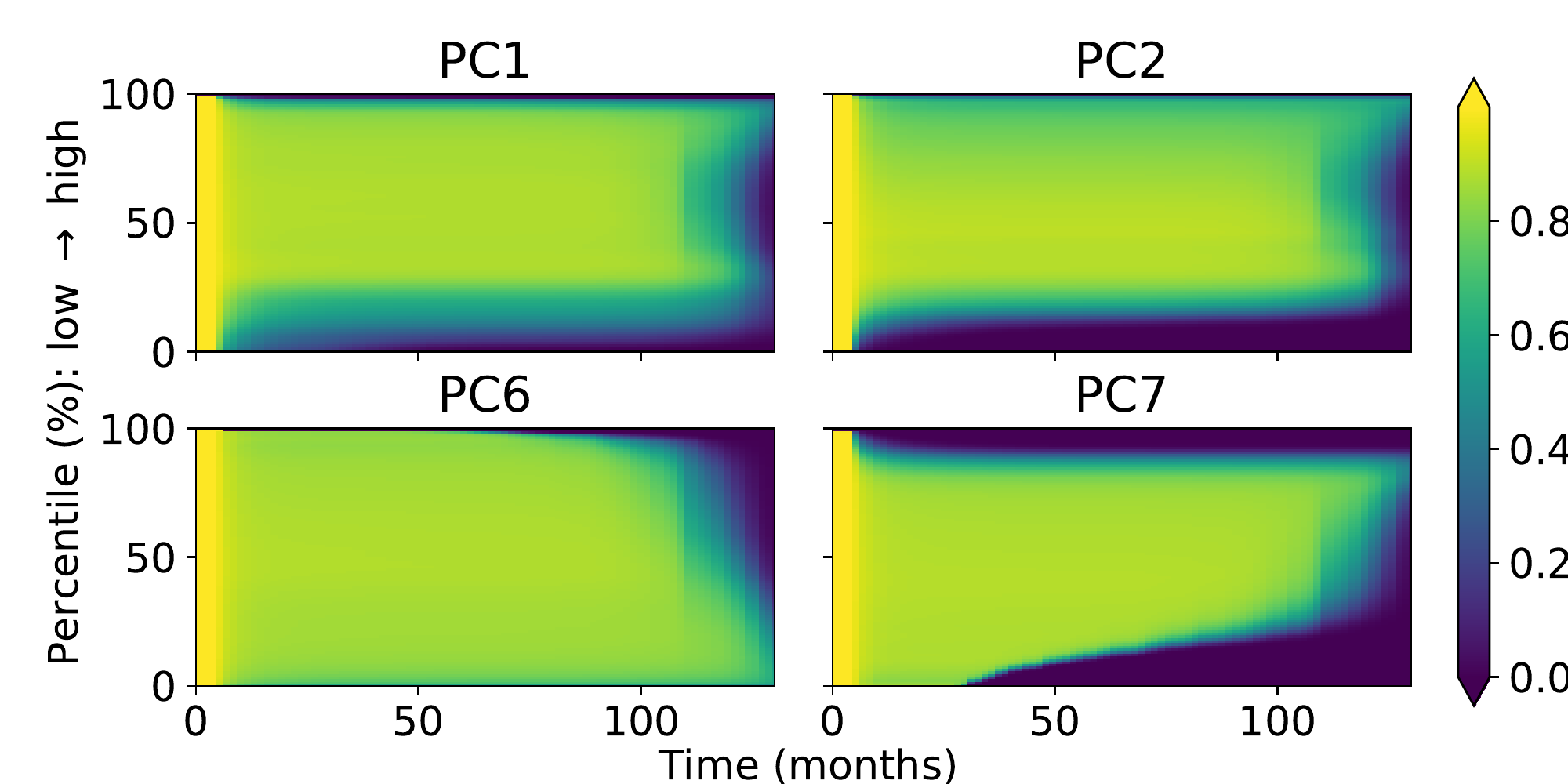}
    \caption{The contribution of predictors to the survival probability $S(t|{\bf x})$ with remaining features fixed. Brighter regions indicate high survival rates. }
    \label{fig:density_sur}
\end{figure}

Results in Table~\ref{table: survival} show that structural factors of the community in which a neologism is introduced can predict its chance of survival or death, with all models outperforming the baseline by a significant margin. 
Since samples in training and test sets do not overlap in subreddits, such performance indicates that there are strong associations between network structures and word survival such that our models can generalize across communities. The coefficients for the Cox model with 10 PCs are shown in Table~\ref{tab:Cox_results}. To interpret the LH model with 10 PCs, we generate the survival function $S(t|x)$ by varying a single feature from low to high but keep the remainder fixed at their median value (Figure~\ref{fig:density_sur}).  While the Cox model predicts the hazard (death rate) and the LH model predicts $S(t|x)$ (the survival rate) (in reverse direction), we found that both models were highly consistent in assessing the input PCs, both in terms of relative weights and directions. 

A large overall size (PC1) tends to preserve neologisms, as large communities provide a basic threshold population for words to be used. In addition to sheer size, global network topology also contributes to neologism survival. PC2, PC3, PC6 and PC7 correspond to three different network structures. PC3 represents networks that have many external connections but are split into multiple clusters within the community, which contributes negatively to the survival probabilities. In contrast, less clustered networks with dense edges and rich external connections (PC2) increase word survival rates. Both PC6 and PC7 boost word survival rate and they both represent networks that are relatively densely connected, but PC6 has high connections to many external communities and is more fragmented whereas PC7 is more isolated in the inter-community network (low degree centrality) but its external connections are influential communities (high Pagerank and Betweenness centrality). This may suggest that inter- and intra-community connections complement each other. In general, within a community, dense connections in the network keep words alive whereas local clusters in the network are adverse to word survival. In the multi-community landscape, more external connections tend to promote word survival. 


\begin{table}[tb]
\small
\centering
\begin{tabular}{lrrr}
{\bf Variables} & {\bf Coef.}  & {\bf Exp(coef)} & {\bf S.E.}  \\\hline
PC1       & -0.122$^{***}$ & 0.885     & 0.002 \\
PC2       & -0.072$^{***}$ & 0.930     & 0.002 \\
PC3       & 0.170$^{***}$ & 1.186     & 0.003 \\
PC4       & 0.009$^{***}$  & 1.001      & 0.001 \\
PC5       & -0.017$^{***}$ & 0.984     & 0.001 \\
PC6       & -0.160$^{***}$ & 0.852     & 0.001 \\
PC7       & -0.516$^{***}$ & 1.675     & 0.002 \\
PC8       & -0.048$^{***}$ & 0.953     & 0.001 \\
PC9       & -0.004$^{***}$ & 1.004     & 0.001 \\
PC10      & -0.054$^{***}$ & 0.947     & 0.002 \\ 
\end{tabular}
\caption{Results of the Cox model. All coefficients are highly significant. Exp(coef) refers to the hazard, or the probability of death. Lower Exp(coef) suggests that this variable is protective. S.E. refers to the standard error of the regression coefficients.}
\label{tab:Cox_results}
\end{table}

\section{Lexical levelling}
{\it Levelling} refers to the gradual replacement of localized linguistic features ({\it marked}) by mainstream linguistic features ({\it unmarked}) over the whole community \cite{kerswill2003dialect}, which has been observed in a wide range of offline linguistic communities due to increasing mobility and external contacts \cite{milroy2002introduction,kerswill2003dialect}. 

\begin{figure*}[tbh]
    \centering
    \includegraphics[width=\columnwidth]{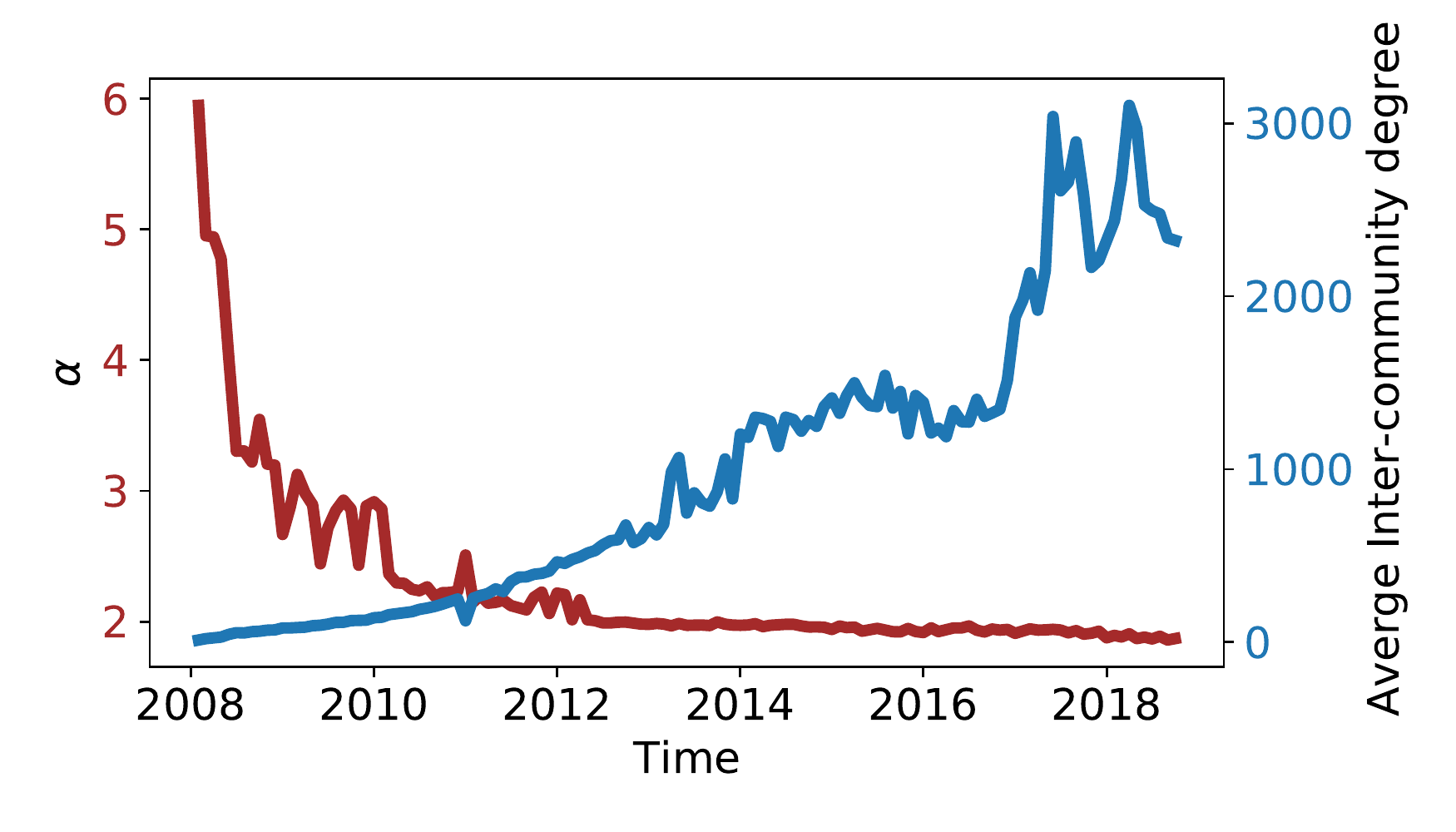}
    \includegraphics[width=\columnwidth]{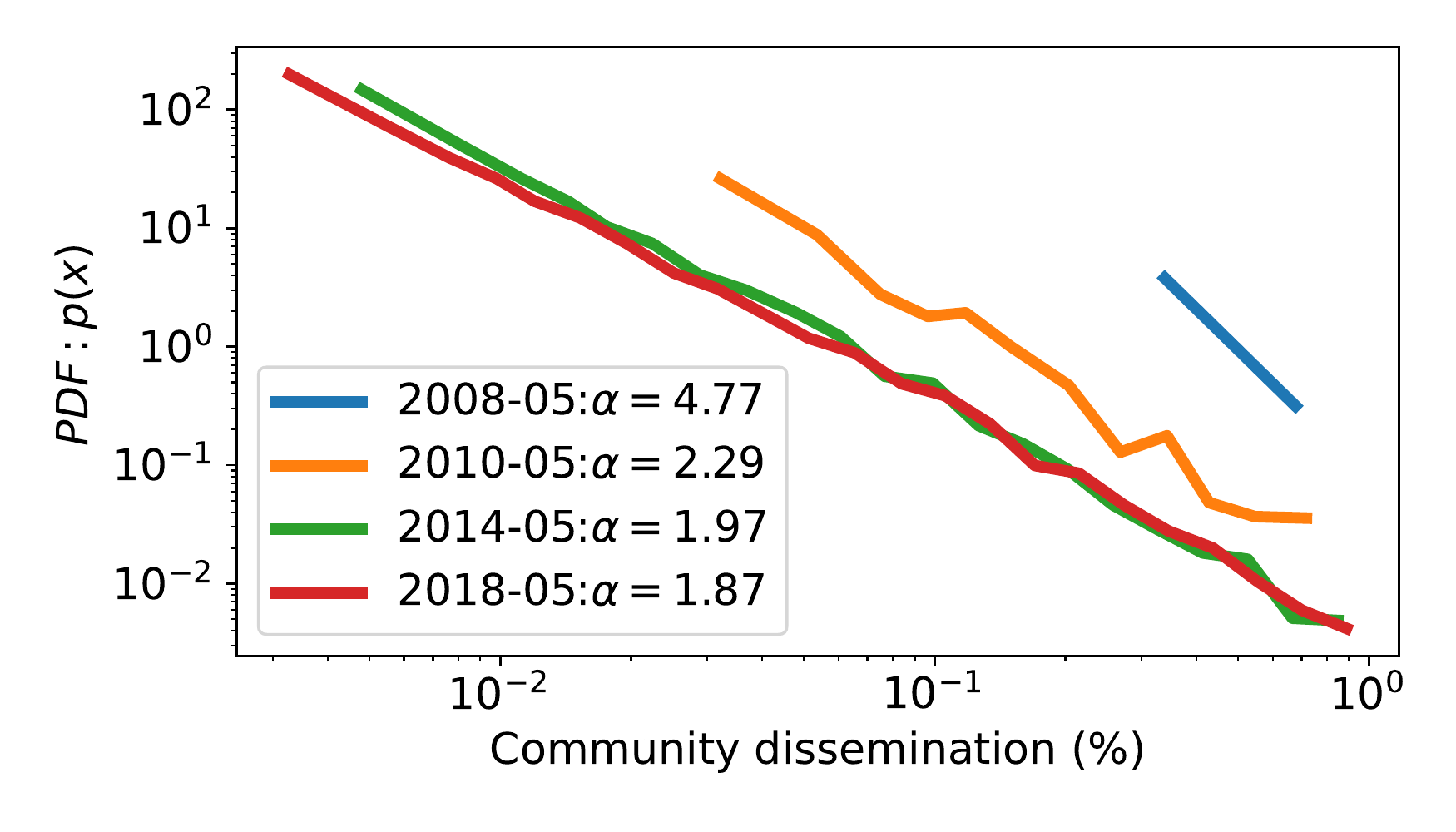}
    \caption{[\textbf{Left}] Change of average community degree and the shape parameter $\alpha$ of the power law fit $p(x) \propto x^{-\alpha}$ over time. The average community degree is increasing, indicating that more communities are connected to each other. The $\alpha$ is decreasing, suggesting that the tail of the distribution has become thicker, or more community specific words have emerged.   [\textbf{Right}] Snapshots of PDFs of dissemination across communities over time.}
    \label{fig:alpha_and_degree}
\end{figure*}

The subreddit communities have become increasingly inter-connected over time, as the average inter-community degree has increased from 6 in January 2008 to 2,323 in October 2018 (Figure~\ref{fig:alpha_and_degree}). While some of these could be accounted for by the simultaneous growth in the number of subreddits, the growth in connectedness is also apparent. Such an increase of contact could promote the spread of neologisms across Reddit. In the same period, the number of variants that spread to more than 60\% of the communities has grown slightly from 7 to 22. Some of the notable examples include words like {\it lol}, {\it alot}, {\it imao} and {\it cuz}. Meanwhile, the variants that are only confined to one community grew rapidly from 1992 in 2008 to 23,397 in 2018. 
The widespread use of some neologisms does not necessarily cause the loss of local expressions, as in offline communities. Instead, the community-specific terms and community-general terms develop in tandem. 
Many community-specific terms are nested within topic-based communities with little meaning overlap with those widespread variants, and are therefore unlikely to be replaced by more general terms through levelling.

Figure~\ref{fig:alpha_and_degree} also shows that the probabilistic density distribution (PDF) of word dissemination (the percentage of communities sharing a neologism) conforms to the power law fit $p(x) \propto x^{-\alpha}$, as a few words spread to most communities while most words are confined to a few communities. Further, the shape parameter $\alpha$ decreases asymptotically despite the growth of average inter-community degree (Figure~\ref{fig:alpha_and_degree}), which implies that, as the size of Reddit grows, more community-specific words, as well as more widespread words, emerge.  

\paragraph{Summary}The number of community specific words grew rapidly despite increased inter-community connectedness, which seems to go against the levelling trend observed in offline networks \cite{conde201218}. In contrast to offline communities, these subreddit networks are of a different nature, as they are topic-based groups bounded by common interests. By joining these communities, users opt for fragmentation into some niche groups. Such segregation in topics and interests naturally brings in more community specific words.
In other words, there is no strong evidence for lexical levelling; instead, online communities go in the reverse direction, by developing more niche neologisms.

\section{Discussions and Conclusions}

In traditional sociolinguistics, weak ties within a social network have been linked to innovation and language change.
Yet most studies only use indirect evidence to infer the underlying network types \cite{milroy1985linguistic,nevalainen2000mobility,dodsworth2019bipartite}. 
Our quantitative analysis suggests that multiple structural properties  play a role in lexical change. The overall network size is the most prominent factor in lexical innovation and survival, as large communities provide the base population to create and use those neologisms. The effect of network size has also been emphasized in other network studies of language \cite{reali2018simpler,raviv2019larger,laitinen2020size}. However, sheer size is only part of the story, as  dense edges between users, the lack of separate local clusters, and rich external connections also promote both lexical innovation and survival. Dense connections within and across communities increase the visibility of neologisms so that they can be imitated by other users, as exposure alone predicts users' information spreading behavior \cite{bakshy2012role}. In contrast, local clustering tends to separate networks into disconnected parts, slowing the spread of new words. These structural attributes are found to facilitate information spread in online social networks \cite{lerman2010information}. On a broader scale, our results suggest that the lexical change process in online social networks may be similar to other information spread processes \cite{guille2013information}.

Our results show that conclusions drawn from offline communities might be insufficient to account for behavior seen in online social networks. While the classic weak tie model emphasizes the role of loose social networks in language change \cite{milroy1985linguistic,nevalainen2000mobility} and has been confirmed in online communities \cite{del-tredici-fernandez-2018-road}, our work further extends this model by showing that a variety of network structural attributes also play a role in language change. Our quantitative analysis also suggests a different leveling process in online communities with implications 
for sociolinguistic theories.

\paragraph{Limitations and future work} One limitation of this study is that topical variation is not explored in depth, because we aimed to look at the contributes of networks alone by smoothing out topical variation with diverse communities. Yet topics have been found to affect users' posting behavior in online communities \cite{mathew2019deep} and niche topics do affect word retention \cite{altmann2011niche}. In Reddit, communities involving certain niche or foreign topics, such as {\tt r/pokemon}, might inherently introduce more lexical innovations than others. Secondly, we only focus on Internet neologisms in Reddit. 
How these neologisms propagate across multiple social media platforms 
and how online and offline neologisms interact remain important questions to be addressed. Thirdly, while our study reveals the general patterns of lexical change, there are multiple sub-categories of neologisms such as discourse markers and name entities. It is of interest to ask whether different sub-categories may exhibit different patterns of usage in online communities. 
These research questions are worth exploring in future work.

\section{Ethical concerns}
In terms of ethical concerns, a great number of low frequency neologisms collected from \texttt{Urban Dictionary} may be considered offensive to specific groups of populations. We collected the word usage data as they were in order to recover as realistic of a lexical landscape in Reddit as possible. However, these offensive words by no means reflect our values. Nor do we endorse the use of these words.

\section*{Acknowledgements}
We thank Professor Patrice Beddor, Professor Will Styler, Julia Mendelsohn, Jiaxin Pei, Zuoyu Tian and Allison Lahnala for their comments on earlier versions of this draft. We are also grateful to all anonymous reviewers for their insightful comments, which helped improve this manuscript greatly.  This material is based upon work supported by the National Science Foundation under Grant No 185022.

\bibliography{anthology,custom}

\begin{thebibliography}{52}
\expandafter\ifx\csname natexlab\endcsname\relax\def\natexlab#1{#1}\fi

\bibitem[{Altmann et~al.(2011)Altmann, Pierrehumbert, and
  Motter}]{altmann2011niche}
Eduardo~G Altmann, Janet~B Pierrehumbert, and Adilson~E Motter. 2011.
\newblock Niche as a determinant of word fate in online groups.
\newblock \emph{PloS one}, 6(5):e19009.

\bibitem[{Antolini et~al.(2005)Antolini, Boracchi, and
  Biganzoli}]{antolini2005time}
Laura Antolini, Patrizia Boracchi, and Elia Biganzoli. 2005.
\newblock A time-dependent discrimination index for survival data.
\newblock \emph{Statistics in Medicine}, 24(24):3927--3944.

\bibitem[{Backstrom et~al.(2012)Backstrom, Boldi, Rosa, Ugander, and
  Vigna}]{backstrom2012four}
Lars Backstrom, Paolo Boldi, Marco Rosa, Johan Ugander, and Sebastiano Vigna.
  2012.
\newblock Four degrees of separation.
\newblock In \emph{Proceedings of the 4th Annual ACM Web Science Conference},
  pages 33--42.

\bibitem[{Bakshy et~al.(2012)Bakshy, Rosenn, Marlow, and
  Adamic}]{bakshy2012role}
Eytan Bakshy, Itamar Rosenn, Cameron Marlow, and Lada Adamic. 2012.
\newblock The role of social networks in information diffusion.
\newblock In \emph{Proceedings of the 21st international conference on World
  Wide Web}, pages 519--528.

\bibitem[{Barnhart(2007)}]{barnhart2007calculus}
David~K Barnhart. 2007.
\newblock A calculus for new words.
\newblock \emph{Dictionaries: Journal of the Dictionary Society of North
  America}, 28(1):132--138.

\bibitem[{Bhandari and Armstrong(2019)}]{bhandari-armstrong-2019-tkol}
Abhinav Bhandari and Caitrin Armstrong. 2019.
\newblock \href {https://doi.org/10.18653/v1/D19-5508} {Tkol, httt, and
  r/radiohead: High affinity terms in {R}eddit communities}.
\newblock In \emph{Proceedings of the 5th Workshop on Noisy User-generated Text
  (W-NUT 2019)}, pages 57--67, Hong Kong, China. Association for Computational
  Linguistics.

\bibitem[{Chang and Danescu-Niculescu-Mizil(2019)}]{chang2019trajectories}
Jonathan Chang and Cristian Danescu-Niculescu-Mizil. 2019.
\newblock Trajectories of blocked community members: Redemption, recidivism and
  departure.
\newblock In \emph{The World Wide Web Conference}, pages 184--195.

\bibitem[{Chang et~al.(2020)Chang, Chiam, Fu, Wang, Zhang, and
  Danescu-Niculescu-Mizil}]{chang2020convokit}
Jonathan~P Chang, Caleb Chiam, Liye Fu, Andrew~Z Wang, Justine Zhang, and
  Cristian Danescu-Niculescu-Mizil. 2020.
\newblock Convokit: A toolkit for the analysis of conversations.
\newblock In \emph{Proceedings of SIGDIAL}.

\bibitem[{Christiansen and Kirby(2003)}]{christiansen2003language}
Morten~H Christiansen and Simon Kirby. 2003.
\newblock Language evolution: Consensus and controversies.
\newblock \emph{Trends in cognitive sciences}, 7(7):300--307.

\bibitem[{Conde-Silvestre(2012)}]{conde201218}
Juan~Camilo Conde-Silvestre. 2012.
\newblock The role of social networks and mobility in diachronic
  sociolinguistics.
\newblock \emph{The Handbook of Historical Sociolinguistics}, pages 332--352.

\bibitem[{Cunha et~al.(2019)Cunha, Jurgens, Tan, and Romero}]{cunha2019all}
Tiago Cunha, David Jurgens, Chenhao Tan, and Daniel Romero. 2019.
\newblock Are all successful communities alike? characterizing and predicting
  the success of online communities.
\newblock In \emph{The World Wide Web Conference}, pages 318--328.

\bibitem[{Danescu-Niculescu-Mizil et~al.(2013)Danescu-Niculescu-Mizil, West,
  Jurafsky, Leskovec, and Potts}]{danescu2013no}
Cristian Danescu-Niculescu-Mizil, Robert West, Dan Jurafsky, Jure Leskovec, and
  Christopher Potts. 2013.
\newblock No country for old members: User lifecycle and linguistic change in
  online communities.
\newblock In \emph{Proceedings of the 22nd international conference on World
  Wide Web}, pages 307--318. ACM.

\bibitem[{Del~Tredici and
  Fern{\'a}ndez(2018)}]{del-tredici-fernandez-2018-road}
Marco Del~Tredici and Raquel Fern{\'a}ndez. 2018.
\newblock \href {https://www.aclweb.org/anthology/C18-1135} {The road to
  success: Assessing the fate of linguistic innovations in online communities}.
\newblock In \emph{Proceedings of the 27th International Conference on
  Computational Linguistics}, pages 1591--1603, Santa Fe, New Mexico, USA.
  Association for Computational Linguistics.

\bibitem[{Dodsworth(2019)}]{dodsworth2019bipartite}
Robin Dodsworth. 2019.
\newblock Bipartite network structures and individual differences in sound
  change.
\newblock \emph{Glossa: a journal of general linguistics}, 4(1).

\bibitem[{Eckert and McConnell-Ginet(1992)}]{eckert1992think}
Penelope Eckert and Sally McConnell-Ginet. 1992.
\newblock Think practically and look locally: Language and gender as
  community-based practice.
\newblock \emph{Annual review of anthropology}, 21(1):461--488.

\bibitem[{Eisenstein et~al.(2014)Eisenstein, O'Connor, Smith, and
  Xing}]{eisenstein2014diffusion}
Jacob Eisenstein, Brendan O'Connor, Noah~A Smith, and Eric~P Xing. 2014.
\newblock Diffusion of lexical change in social media.
\newblock \emph{PloS one}, 9(11).

\bibitem[{Goel et~al.(2016)Goel, Soni, Goyal, Paparrizos, Wallach, Diaz, and
  Eisenstein}]{goel2016social}
Rahul Goel, Sandeep Soni, Naman Goyal, John Paparrizos, Hanna Wallach, Fernando
  Diaz, and Jacob Eisenstein. 2016.
\newblock The social dynamics of language change in online networks.
\newblock In \emph{International Conference on Social Informatics}, pages
  41--57. Springer.

\bibitem[{Granovetter(1983)}]{granovetter1983strength}
Mark Granovetter. 1983.
\newblock The strength of weak ties: A network theory revisited.
\newblock \emph{Sociological theory}, pages 201--233.

\bibitem[{Granovetter(1977)}]{granovetter1977strength}
Mark~S Granovetter. 1977.
\newblock The strength of weak ties.
\newblock In \emph{Social networks}, pages 347--367. Elsevier.

\bibitem[{Grieve et~al.(2017)Grieve, Nini, and Guo}]{grieve2017analyzing}
Jack Grieve, Andrea Nini, and Diansheng Guo. 2017.
\newblock Analyzing lexical emergence in modern american english online 1.
\newblock \emph{English Language \& Linguistics}, 21(1):99--127.

\bibitem[{Guille et~al.(2013)Guille, Hacid, Favre, and
  Zighed}]{guille2013information}
Adrien Guille, Hakim Hacid, Cecile Favre, and Djamel~A Zighed. 2013.
\newblock Information diffusion in online social networks: A survey.
\newblock \emph{ACM Sigmod Record}, 42(2):17--28.

\bibitem[{Hamilton et~al.(2017)Hamilton, Zhang, Danescu-Niculescu-Mizil,
  Jurafsky, and Leskovec}]{hamilton2017loyalty}
William~L Hamilton, Justine Zhang, Cristian Danescu-Niculescu-Mizil, Dan
  Jurafsky, and Jure Leskovec. 2017.
\newblock Loyalty in online communities.
\newblock In \emph{Eleventh International AAAI Conference on Web and Social
  Media}.

\bibitem[{Holmes and Meyerhoff(1999)}]{holmes1999community}
Janet Holmes and Miriam Meyerhoff. 1999.
\newblock The community of practice: Theories and methodologies in language and
  gender research.
\newblock \emph{Language in society}, 28(2):173--183.

\bibitem[{Kershaw et~al.(2016)Kershaw, Rowe, and Stacey}]{kershaw2016towards}
Daniel Kershaw, Matthew Rowe, and Patrick Stacey. 2016.
\newblock Towards modelling language innovation acceptance in online social
  networks.
\newblock In \emph{Proceedings of the Ninth ACM International Conference on Web
  Search and Data Mining}, pages 553--562.

\bibitem[{Kerswill(2003)}]{kerswill2003dialect}
Paul Kerswill. 2003.
\newblock Dialect levelling and geographical diffusion in british english.
\newblock \emph{Social dialectology: in honour of Peter Trudgill}, pages
  223--243.

\bibitem[{Klein et~al.(2019)Klein, Clutton, and Dunn}]{klein2019pathways}
Colin Klein, Peter Clutton, and Adam~G Dunn. 2019.
\newblock Pathways to conspiracy: The social and linguistic precursors of
  involvement in reddit’s conspiracy theory forum.
\newblock \emph{PloS one}, 14(11):e0225098.

\bibitem[{Kleinbaum and Klein(2010)}]{kleinbaum2010survival}
David~G Kleinbaum and Mitchel Klein. 2010.
\newblock \emph{Survival analysis}.
\newblock Springer.

\bibitem[{Kvamme et~al.(2019)Kvamme, {{\O}}rnulf Borgan, and
  Scheel}]{JMLR:v20:18-424}
H{{\aa}}vard Kvamme, {{\O}}rnulf Borgan, and Ida Scheel. 2019.
\newblock \href {http://jmlr.org/papers/v20/18-424.html} {Time-to-event
  prediction with neural networks and cox regression}.
\newblock \emph{Journal of Machine Learning Research}, 20(129):1--30.

\bibitem[{Kvamme and Borgan(2019)}]{kvamme2019continuous}
H{\aa}vard Kvamme and {\O}rnulf Borgan. 2019.
\newblock Continuous and discrete-time survival prediction with neural
  networks.
\newblock \emph{arXiv preprint arXiv:1910.06724}.

\bibitem[{Labov(2007)}]{labov2007transmission}
William Labov. 2007.
\newblock Transmission and diffusion.
\newblock \emph{Language}, 83(2):344--387.

\bibitem[{Laitinen et~al.(2020)Laitinen, Fatemi, and
  Lundberg}]{laitinen2020size}
Mikko Laitinen, Masoud Fatemi, and Jonas Lundberg. 2020.
\newblock Size matters: Digital social networks and language change.
\newblock \emph{Frontiers in Artificial Intelligence}, 3:46.

\bibitem[{Lerman and Ghosh(2010)}]{lerman2010information}
Kristina Lerman and Rumi Ghosh. 2010.
\newblock Information contagion: An empirical study of the spread of news on
  {D}igg and {T}witter social networks.
\newblock In \emph{Proceedings of 4th International Conference on Weblogs and
  Social Media (ICWSM), 2010}.

\bibitem[{Lupyan and Dale(2010)}]{lupyan2010language}
Gary Lupyan and Rick Dale. 2010.
\newblock Language structure is partly determined by social structure.
\newblock \emph{PloS one}, 5(1).

\bibitem[{Mathew et~al.(2019)Mathew, Dutt, Maity, Goyal, and
  Mukherjee}]{mathew2019deep}
Binny Mathew, Ritam Dutt, Suman~Kalyan Maity, Pawan Goyal, and Animesh
  Mukherjee. 2019.
\newblock Deep dive into anonymity: Large scale analysis of {Q}uora questions.
\newblock In \emph{International Conference on Social Informatics}, pages
  35--49. Springer.

\bibitem[{Metcalf(2004)}]{metcalf2004predicting}
Allan~A Metcalf. 2004.
\newblock \emph{Predicting new words: The secrets of their success}.
\newblock Houghton Mifflin Harcourt.

\bibitem[{Milroy and Milroy(1985)}]{milroy1985linguistic}
James Milroy and Lesley Milroy. 1985.
\newblock Linguistic change, social network and speaker innovation.
\newblock \emph{Journal of Linguistics}, 21(2):339--384.

\bibitem[{Milroy(2002)}]{milroy2002introduction}
Lesley Milroy. 2002.
\newblock Introduction: Mobility, contact, and language change--working with
  contemporary speech communities.
\newblock \emph{Journal of Sociolinguistics}, 6(1):3--15.

\bibitem[{Milroy and Llamas(2013)}]{milroy2013social}
Lesley Milroy and Carmen Llamas. 2013.
\newblock Social networks.
\newblock \emph{The Handbook of Language Variation and Change}, pages 407--427.

\bibitem[{Milroy and Milroy(1992)}]{milroy1992social}
Lesley Milroy and James Milroy. 1992.
\newblock Social network and social class: Toward an integrated sociolinguistic
  model.
\newblock \emph{Language in society}, 21(1):1--26.

\bibitem[{Nevalainen(2000)}]{nevalainen2000mobility}
Terttu Nevalainen. 2000.
\newblock Mobility, social networks and language change in {E}arly {M}odern
  {E}ngland.
\newblock \emph{European Journal of English Studies}, 4(3):253--264.

\bibitem[{Newberry et~al.(2017)Newberry, Ahern, Clark, and
  Plotkin}]{newberry2017detecting}
Mitchell~G Newberry, Christopher~A Ahern, Robin Clark, and Joshua~B Plotkin.
  2017.
\newblock Detecting evolutionary forces in language change.
\newblock \emph{Nature}, 551(7679):223--226.

\bibitem[{Noble and Fern{\'a}ndez(2015)}]{noble-fernandez-2015-centre}
Bill Noble and Raquel Fern{\'a}ndez. 2015.
\newblock \href {https://doi.org/10.3115/v1/W15-1104} {Centre stage: How social
  network position shapes linguistic coordination}.
\newblock In \emph{Proceedings of the 6th Workshop on Cognitive Modeling and
  Computational Linguistics}, pages 29--38, Denver, Colorado. Association for
  Computational Linguistics.

\bibitem[{Page et~al.(1999)Page, Brin, Motwani, and
  Winograd}]{page1999pagerank}
Lawrence Page, Sergey Brin, Rajeev Motwani, and Terry Winograd. 1999.
\newblock The pagerank citation ranking: Bringing order to the web.
\newblock Technical report, Stanford InfoLab.

\bibitem[{Paolillo(1999)}]{paolillo1999virtual}
John~C Paolillo. 1999.
\newblock The virtual speech community: Social network and language variation
  on irc.
\newblock In \emph{Proceedings of the 32nd Annual Hawaii International
  Conference on Systems Sciences. 1999. HICSS-32. Abstracts and CD-ROM of Full
  Papers}, pages 10--pp. IEEE.

\bibitem[{Paradowski and Jonak(2012)}]{paradowski2012diffusion}
Micha{\l}~B Paradowski and {\L}ukasz Jonak. 2012.
\newblock Diffusion of linguistic innovation as social coordination.
\newblock \emph{Psychology of Language and Communication}, 16(2):131--142.

\bibitem[{Raviv et~al.(2019)Raviv, Meyer, and Lev-Ari}]{raviv2019larger}
Limor Raviv, Antje Meyer, and Shiri Lev-Ari. 2019.
\newblock Larger communities create more systematic languages.
\newblock \emph{Proceedings of the Royal Society B}, 286(1907):20191262.

\bibitem[{Reali et~al.(2018)Reali, Chater, and Christiansen}]{reali2018simpler}
Florencia Reali, Nick Chater, and Morten~H Christiansen. 2018.
\newblock Simpler grammar, larger vocabulary: How population size affects
  language.
\newblock \emph{Proceedings of the Royal Society B: Biological Sciences},
  285(1871):20172586.

\bibitem[{Schwen and Hara(2003)}]{schwen2003community}
Thomas~M Schwen and Noriko Hara. 2003.
\newblock Community of practice: A metaphor for online design?
\newblock \emph{The Information Society}, 19(3):257--270.

\bibitem[{Sharma and Dodsworth(2020)}]{sharma2020language}
Devyani Sharma and Robin Dodsworth. 2020.
\newblock Language variation and social networks.
\newblock \emph{Annual Review of Linguistics}, 6.

\bibitem[{Stewart and Eisenstein(2018)}]{stewart-eisenstein-2018-making}
Ian Stewart and Jacob Eisenstein. 2018.
\newblock \href {https://doi.org/10.18653/v1/D18-1467} {Making {``}fetch{''}
  happen: The influence of social and linguistic context on nonstandard word
  growth and decline}.
\newblock In \emph{Proceedings of the 2018 Conference on Empirical Methods in
  Natural Language Processing}, pages 4360--4370, Brussels, Belgium.
  Association for Computational Linguistics.

\bibitem[{Tan and Lee(2015)}]{tan2015all}
Chenhao Tan and Lillian Lee. 2015.
\newblock All who wander: On the prevalence and characteristics of
  multi-community engagement.
\newblock In \emph{Proceedings of the 24th International Conference on World
  Wide Web}, pages 1056--1066.

\bibitem[{Zhang et~al.(2017)Zhang, Hamilton, Danescu-Niculescu-Mizil, Jurafsky,
  and Leskovec}]{zhang2017community}
Justine Zhang, William Hamilton, Cristian Danescu-Niculescu-Mizil, Dan
  Jurafsky, and Jure Leskovec. 2017.
\newblock Community identity and user engagement in a multi-community
  landscape.
\newblock In \emph{Proceedings of the International AAAI Conference on Web and
  Social Media}, volume~11.

\end{thebibliography}
\bibliographystyle{acl_natbib}

\clearpage
\newpage
\appendix
\section{Replicability}\label{replic}
We take measures to ensure the replicability of our study. Some of the validation results are presented in the following supplementary materials. 
The following resources can be used to replicate the current study.
\begin{itemize}
    \item Our code for preprocessing and analysis as well as the preprocessed data can be found at: \\
    \url{https://github.com/lingjzhu/reddit_network}.
    \item The original Reddit data can be retrieved from the following corpus (about 2TB): \\
    \url{https://convokit.cornell.edu/documentation/subreddit.html}.
    \item The list of neologisms was collected from the \texttt{Urban Dictionary} and \texttt{NoSlang.com}. {\it Warning: the following two sites may contain offensive content.} \\
    \url{https://www.urbandictionary.com/} \\
    \url{https://www.noslang.com/}.
    \item The network construction was carried out using the \texttt{networkx} while the feature extraction was through the \texttt{networkit} packages: \\
    \url{https://networkx.org/} \\
    \url{https://networkit.github.io/}.
    \item The statistical tests were implemented with {\tt Pingouin} and predictive models with {\tt sklearn}. \\
    \url{https://pingouin-stats.org/} \\
    \url{https://scikit-learn.org/stable/}
    \item The deep survival analysis was implemented using the \texttt{pycox} package: \\
    \url{https://github.com/havakv/pycox}.
    \item The baseline cox model was implemented using the \texttt{lifelines} package: \\
    \url{https://lifelines.readthedocs.io/en/latest/Survival\%20Regression.html}.
\end{itemize}

\section{The Reddit Network Corpus}
\label{app: corpus}

The detailed information of the Reddit Network Corpus is given in this section. The code and data will soon be released to the public.

\subsection{Neologisms}
Table~\ref{table:exp_neo} shows some samples of most frequent and least frequent neologisms in Reddits. These linguistic innovations were collected from \texttt{NoSlang.com} and \texttt{Urban Dictionary}. We filtered out lexical entries that: 1) span more than one word, 2) can be found as an entry in an English dictionary after lemmatization, 3) are identified as person names, 4) contain non-alphabetical characters, numbers or emojis and 5) do not show up in our Reddit dataset.

We set loose criteria for word inclusion. Many of the frequent neologisms have already been incorporated into the daily lexicon, such as {\it wiki}, {\it google} and {\it instagram}. We manually filtered out these words in our wordlist and the number of such words is less than 100.   
We also keep typos in the curated list, as these words often carry special meanings. For example, {\it alot}, {\it atleast} and {\it recieve} are the typos that are used more than 1 million times, so frequent that they carry some special meanings and functions such as identity assertion. 

After automatic filtering, we manually inspected the 5000 most frequent words with greater care so as to filter out some invalid entries. In addition, we also sampled a few hundred words at different frequency bins for close inspection. For the rest of the words, we only scanned through them for a quick sanity check.


\section{Additional validation of the networks}\label{app: network}
\subsection{Intra-community networks}
We constructed the network representations of Reddit communities with the same method as that used by \citet{hamilton2017loyalty} and \citet{del-tredici-fernandez-2018-road}, so that our study is consistent and comparable with previous works. The rationale behind this setting is that "two users who comment in such
proximity interacted with each other, or at least directly with
the same material" \cite{hamilton2017loyalty}. 

Here we compare the inter-community networks in our study with two types of baseline networks extracted from the same Reddit communities. We randomly sampled 100 networks from our data and created the following two baseline networks. 
\begin{itemize}
    \item {\bf DRG}: The Direct-reply Graph (DRG) was constructed by treating every user as a node. An edge was created between two users if one user directly replied to the other. This network could {\it underestimate} the user interactions as users are likely to read nearby posts in the same comment chain when replying.
    
    \item {\bf TG}: The Thread Graph (TG) was constructed by setting each user as a node and two users were connected by an edge if they had commented in the same thread. This network might {\it overestimate} the user interactions because in some mega threads that span hundreds or thousands of posts, users might not interact with all the people in the same thread but only with nearby users. 
\end{itemize}
As these two baseline networks might either underestimate or overestimate the connections, we used these two networks to provide an estimate of the possible errors of our networks. 

The results are presented in Table~\ref{tab: intra-corr}. Despite the different settings, most of the network parameters have correlations ranging from moderate to strong. But the correlations for assortativity and clustering coefficients are weaker. However, TG is not considered a good indication of the connections in Reddit as users are unlikely to interact with all users in a long thread. DRG and our networks are more similar to each other. 
\citet{hamilton2017loyalty} had noted that changing the original networks to DRG did not significantly change their analysis results of Reddit networks.


\begin{table}[]
\begin{tabular}{lll}
\hline
\multicolumn{1}{c}{\multirow{2}{*}{{\bf Variables}}} & \multicolumn{2}{c}{{\bf Kendall $\tau$ correlations}} \\
\multicolumn{1}{c}{}                           & {\bf TG}             & {\bf DRG}             \\\hline
\# Nodes                                       & 1 (0.0)        & 1 (0.0)         \\
\# Edges                                       & 0.81 (0.27)    & 0.957 (0.05)    \\
density                                        & 0.50 (0.34)    & 0.928 (0.08)    \\
assortativity                                  & 0.04 (0.26)    & 0.189 (0.25)    \\
local clustering                               & 0.40 (0.18)    & 0.21 (0.36)     \\
global clustering                              & 0.07 (0.29)    & 0.56 (0.25)     \\
average degree                                 & 0.54 (0.30)    & 0.84 (0.14)     \\
max degree                                     & 0.75 (0.12)    & 0.66 (0.22)     \\
min degree                                     & 0.34 (0.03)    & 0.44 (0.09)     \\
LCC \%                                         & 0.60 (0.19)    & 0.55 (0.19)     \\
Singletons \%                                  & 0.53 (0.22)    & 0.49 (0.22)    \\\hline
\end{tabular}
\caption{Correlations between two baseline intra-community networks and the intra-community networks used in our study. The reported numbers are mean correlations with standard deviations inside the bracket.}
\label{tab: intra-corr}
\end{table}

\subsection{Inter-community networks}
In order to validate our approach to construct the inter-community graph, we constructed different inter-community graphs by setting the posting threshold of active users to 2, 3 and 4. One concern is that setting the threshold too low (>=1) results in extremely dense graphs, which are challenging to process. 

After extracting the network features from these networks, we compared them by computing the Kendall rank correlation coefficients between these features. The results in Table~\ref{tab: inter-corr} show that these networks are highly correlated in structural features, especially for degree, eigenvector and pagerank centralities. The correlations for betweenness and closeness are more unstable but still moderately correlated. So adjusting the threshold does not significantly bias our results qualitatively. 


\begin{table}[]
\begin{tabular}{lcc}
\hline
\multirow{2}{*}{{\bf Centrality}} & \multicolumn{2}{c}{{\bf Kendall $\tau$ correlations}} \\
                            & {\bf Threshold: 3}             & {\bf Threshold: 4}             \\\hline
Betweenness                 & 0.62 (0.26)  & 0.48 (0.37)   \\
Closeness                   & 0.64 (0.17)  & 0.50 (0.25)  \\
Degree                      & 0.85 (0.05)   & 0.79 (0.06)  \\
Eigenvector                 & 0.92 (0.03)  & 0.88 (0.05)  \\
Pagerank                    & 0.90 (0.04)  & 0.86 (0.06) \\\hline
\end{tabular}
\caption{Correlations between two baseline inter-community networks and the inter-community networks used in our study. The reported numbers are mean correlations with standard deviations inside the bracket.}
\label{tab: inter-corr}
\end{table}

\subsection{Network statistics}
Table~\ref{table:general_stats} provides some general statistics of the whole Reddit Network Corpus. 

\begin{table}[tbh]
\centering
\begin{tabular}{lr}
\hline  & Total \\ \hline
Months  & 152  \\
Subreddits & 4420  \\
Inter.Networks & 152  \\
Intra.Networks & 289170   \\
Users & $>$ 50 millions\\
Neologisms & 80071  \\
\hline
\end{tabular}
\caption{Summary of the whole corpus.}
\label{table:general_stats}
\end{table}

\begin{table*}[tbh]
\centering
\begin{tabular}{lrr}
\hline  &  Median & Inter-quartile range\\ \hline
Nodes & 765 & [351,  1776]\\
Density & 0.0059  & [0.0025,  0.0121] \\
Average Degree & 4.19 & [3.04,  6.29] \\
Largest Connected Component & 88.6\% & [81.9\%, 92.4\%] \\
Singletons & 9.4\% & [6.5\%, 13.9\%] \\
Inter-Community Degree & 1351 & [492,  2322]   \\
\hline
\end{tabular}
\caption{Statistical Summary of 289170 Subreddit networks.}
\label{table:graph_stats}
\end{table*}

The average duration of the 4420 subreddit communities is 65 months. Statistical summaries of all 289170 networks are presented in Table~\ref{table:graph_stats}.

\section{Correlations between variables}\label{app:corr}
\begin{figure*}
    \centering
    \includegraphics{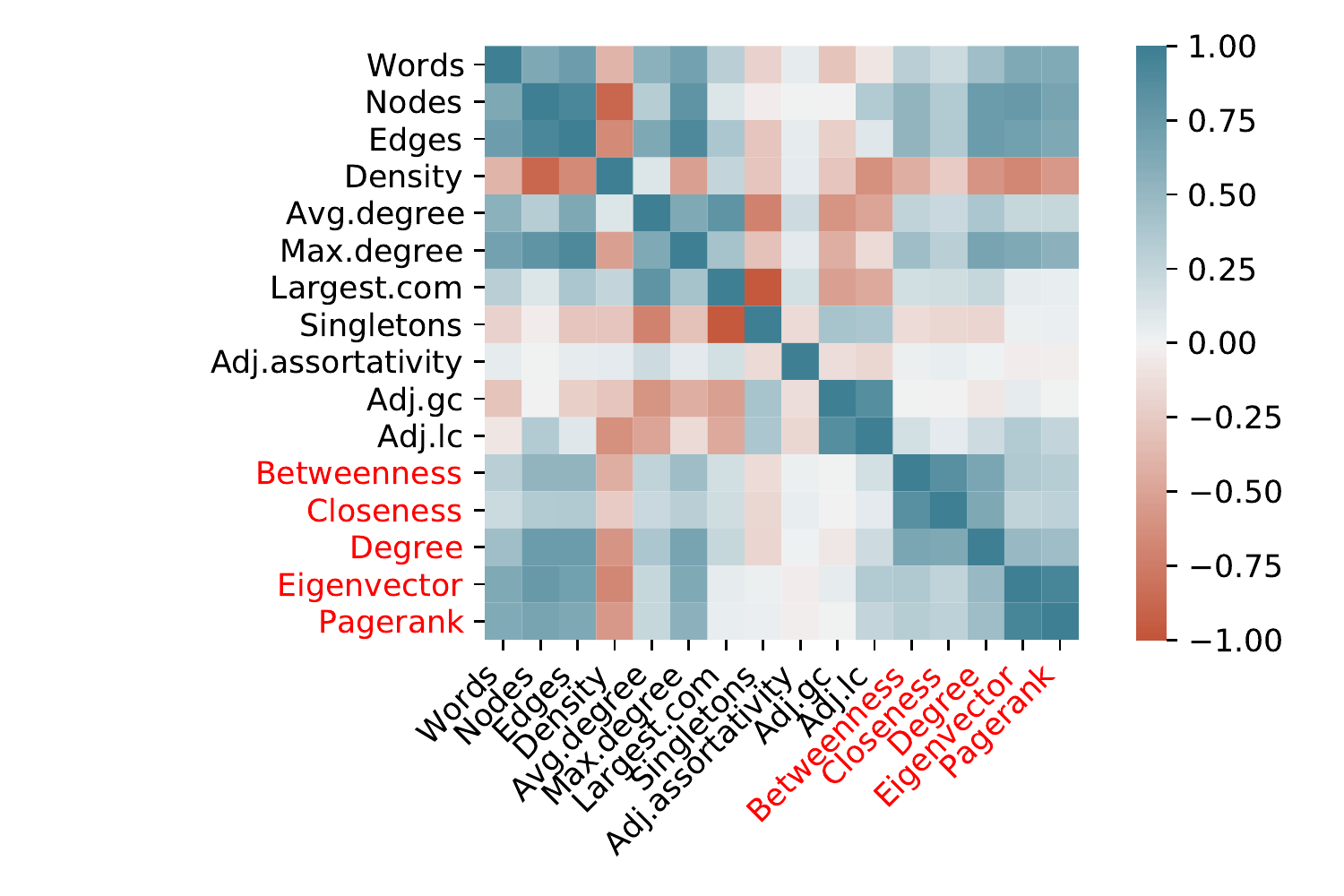}
    \caption{The correlation matrix between variables. Correlation coefficients were computed using the Spearman correlation, as some variables are related in log-linear relations. Variables in dark are within-community features while variables in red are inter-community features.}
    \label{fig:corr}
\end{figure*}

For empirical networks, some network attributes are often correlated. Here we present the correlation matrix between variables used in innovation prediction in Figure~\ref{fig:corr} for illustration. The correlation matrix for features in survival analysis also exhibit a similar pattern of correlations. 

\section{Predicting lexical innovations}
\label{app:innovate}

\subsection{Feature preprocessing}\label{fpro}
We used mean-variance normalization to normalize all prediction features. Since the distribution of some features were highly skewed, before normalization, we log-transformed the following intra-community features: {\it number of nodes, number of edges, density, average degree, maximum degree}, and the following inter-community features: {\it degree centrality, closeness centrality, Pagerank centrality, betweenness centrality, eigenvector centrality}. The rest of the features were directly normalized. Whether to perform log-transformation was determined by visual inspection of the density plot. A small number $10^{-6}$ was added before taking the logarithm to improve numerical stability. We found that such a practice improved the performance during cross-validation relative to directly normalizing all features. 

The following features were used to predict the number of innovations per-month. Some of the features were correlated and the correlations varied from weak to strong. 
\begin{itemize}
    \item {\bf Inter.}: {\it degree centrality, closeness centrality, Pagerank centrality, betweenness centrality, eigenvector centrality}
    \item {\bf Intra.}: {\it number of nodes, number of edges, density, average degree, maximum degree, proportion of the largest connected components, proportion of singletons, adjusted assortativity, adjusted transitivity, adjusted clustering coefficients}.
\end{itemize}

Then PCA with whitening was applied to decompose all of the features into principal components. We did consider the delta features, which were the change in these variables with respect to the last month. However, these added temporal features did not improve the performance. So we assumed that changes in each month might not be highly relevant.

\begin{table}[]
    \centering
    \begin{tabular}{cc}\hline
    Variables  & Coefficients \\\hline
       PC1  &  -0.877\\
       PC2  &  -0.195 \\
       PC3 &  0.193 \\
       PC4 & -0.024 \\
       PC5 & -0.003 \\\hline
    \end{tabular}
    \caption{{\bf [Predicting innovations]} Coefficients for the Poisson regression with first 5 PCs.}
    \label{tab:my_label}
\end{table}

\subsection{Implementation}
All models were implemented in \texttt{sklearn}. The baseline was the mean number of innovations across all time and all subreddits as the prediction. For the rest of models, we performed ten-fold cross-validation to select the best parameters. After parameter selection, the regularization parameter for the Poisson regression was $10^{-2}$ and the maximum number of iterations was 300. For the histogram based gradient boosting trees, the maximum number of split was set to 256 and the loss was the Poisson loss. Otherwise we kept the default hyperparameters.

The data were partitioned into training and test sets with a ratio of 90\%/10\%. We ran each model 20 times with a different random partition each time. The resulting metrics were averaged across the metrics obtained from the test sets over 20 runs.

\begin{figure*}
    \centering
    \includegraphics[width=\textwidth]{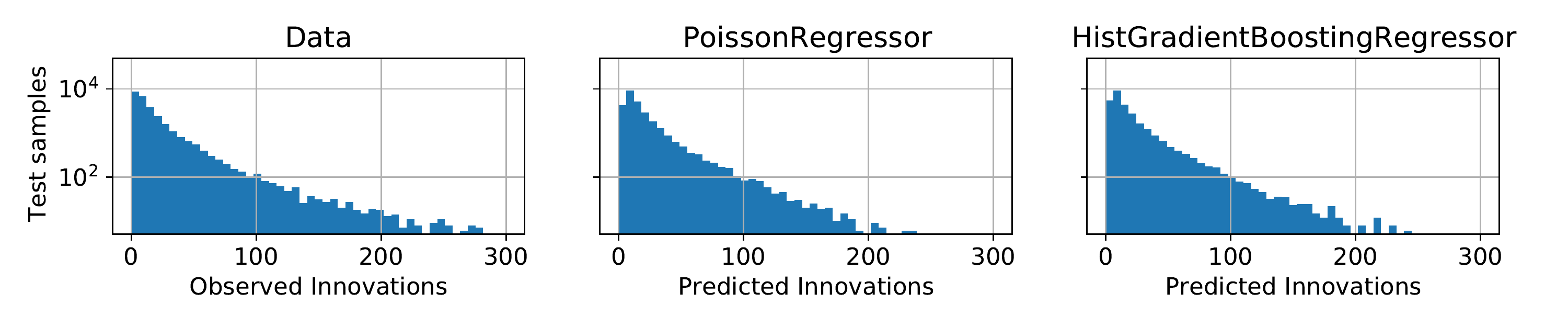}
    \caption{{\bf [Predicting innovations]} Distribution of observed innovations in test samples (left) and predicted distributions by Poisson regression with all network features (middle) and by HGBT with all network features (right). Both models well approximate the empirical distribution of lexical innovation counts but fall short of predicting the trailing long tail.}
    \label{fig:inno_dist}
\end{figure*}

\section{Deep survival analysis}
\label{app: survival}
In this section, we describe the details of deep survival analysis.
\subsection{Model specification}
We adopted the Logistic Hazard model developed in following works \cite{kvamme2019continuous,JMLR:v20:18-424}. The original derivation comes from \citet{kvamme2019continuous}.

In survival analysis, given a set of discrete time steps $T=\{t_1,t_2,\dots,t_n\}$ and the event time $t^*$, the goal is to estimate the probability mass distribution (PMF) of the event time $f(t)$ and the survival function $S(t)$.

\begin{equation}
    \begin{aligned}
    &f(t) = P(t^*=t_i), \\
    &S(t) = P(t^*>t_i) = \sum_{j>i}f(t_j) 
    \end{aligned}
\end{equation}

The model can also be expressed as the hazard function $h(t)$.
\begin{equation}
    \begin{aligned}
    h(t) &= P(t^*=t_i|t^*>t_{i-1}) \\
        &=\frac{f(t_i)}{S(t_{i-1})} \\
        &= \frac{S(t_{i-1})-S(t_i)}{S(t_{i-1})}
    \end{aligned}
\end{equation}

With the above equations, the survival function can be rewritten as follows.
\begin{equation}
    \begin{aligned}
    f(t_i) = h(t_i)S(t_{i-1}) \\
    S(t_i) = [1-h(t_i)]S(t_{i-1})
    \end{aligned}
\end{equation}

It then follows that
\begin{equation}
    S(t_i) = \prod_{k=1}^i[1-h(t_k)]
\end{equation}

For each individual $i$, the likelihood function can be formulated as
\begin{equation}
    L_i = f(t_i)^{d_i}S(t_i)^{1-d_i}
\end{equation}

The above equation can be rewritten with respect to the hazard function.
\begin{equation}
    \begin{aligned}
     L_i =& f(t_i)^{d_i}S(t_i)^{1-d_i} \\
     =&[h(t_i)S(t_{i-1})]^{d_i}\Big([1-h(t_i)]S(t_{i-1})\Big)^{1-d_i} \\
     =& h(t_i)^{d_i}[1-h(t_i)]^{1-d_i}S(t_{i-1}) \\
     =& h(t_i)^{d_i}[1-h(t_i)]^{1-d_i}\prod_{k=1}^{i-1}[1-h(t_k)]
    \end{aligned}
\end{equation}

The loss function is negative log likelihood function, the negative of the sum of  $log(L_i)$ over all samples. After some algebraic operations, the loss function of the Logistic Hazard model can be formulated as the common binary cross-entropy function. 

\begin{equation}\label{eq:loss}
    \begin{aligned}
            \mathcal{L}=-\frac{1}{n}\sum_{i=1}^n\sum_{j=1}^T\big(y_{ij}log[h(t_j|\textbf{x}_i)]\\ +(1-y_{ij})log[1-h(t_j|\textbf{x}_i)]\big)
    \end{aligned}
\end{equation}
where $y_{ij}$ is the binary event indicator for sample $i$ at time $t$.

Let ${\bf x}$ be an input feature vector and $\phi({\bf x})\in \mathbb{R}^h$ is the neural network that transforms input ${\bf x}$ into $h$ output vectors. 
Each output vector corresponds to a discrete time step such that $\phi({\bf x})=\{\phi_1({\bf x}),\phi_2({\bf x}),\ldots,\phi_h({\bf x})\}$. The hazard function then can be approximated by the sigmoid function.

\begin{equation}
    h(t_i|{\bf x}) = \frac{1}{1+exp[-\phi_i({\bf x})]}
\end{equation}

\begin{figure*}[tbh]
    \centering
    \includegraphics[width=\textwidth]{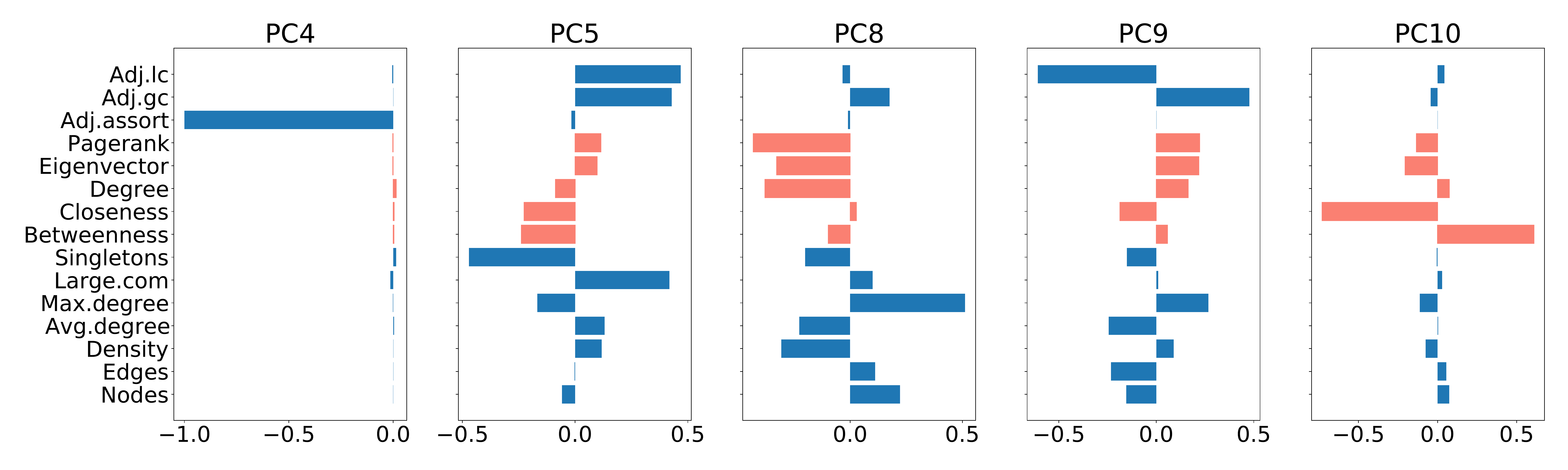}
    \caption{{\bf [Predicting survival]} Remaining PCs for the network features. Inter-community features are highlighted in orange bars. }
    \label{fig:pcs_app}
\end{figure*}

\subsection{Implementation}
Models of deep survival analysis were implemented using the package \texttt{pycox} \cite{JMLR:v20:18-424}. The network features were normalized and partitioned in the same way as described in Section~\ref{fpro}.

The actual survival time for these neologisms varied from 3 to 152 months. First, we discretized the survival time measured in actual months into 100 intervals based on the distribution of the event times, with the assumption that each interval has the same decrease of the survival probability. The resulting grid was denser during months with more event times and sparser during months with fewer event times. Such a practice is recommended by \citet{kvamme2019continuous}, as it reduces parameters and stabilizes training.

We trained a three-layered Logistic Hazard model. For each of the first two layers, we used a linear layer with 256 hidden dimensions and ReLU activation function, followed by batch normalization and a dropout with a probability of 0.1. The last layer was a linear layer with output dimension of 100 followed by a sigmoid activation function.

During training, we used the Adam optimizer with a learning rate of 0.001 and a batch size of 2048 samples. All hyperparameters were tuned with a simple grid search on the development set. Each model was trained for 5 epochs and was run 10 times with different random seeds and different partitions of data each time. The performance metrics were averaged over all 10 runs. These models were trained on a Nvidia V100 GPU and each run took about less than a minute to complete. 

In each run, the data were randomly partitioned into around 80\%, 10\% and 10\% portions as training, development and test sets with different random seeds. In order to avoid information leaking, we ensured that samples in these three sets were from distinct subreddits.

\subsection{Baseline models}
We also ran baseline Cox’s proportional hazard models \cite{kleinbaum2010survival} with the same data partitions and discretization scheme. The Cox's model estimates the hazard function $h(t_i|{\bf x})$ with the following equations.

\begin{equation}
    h(t_i|{\bf x}) = \beta_0(t_i)\cdot \exp\Big(\sum_{i=1}^n\beta_i({\bf x}-\overline{{\bf x}}) \Big)
\end{equation}

We ran the model ten times and report the average performance. All baseline Cox's models were implemented using the \texttt{CoxPHFitter} function via the package \texttt{lifelines}.

\subsection{Additional results of PCA}
The additional results of PCA are shown in Figure~\ref{fig:pcs_app}.

\subsection{Additional results of deep survival analysis}
\begin{figure}[tbh]
    \centering
    \includegraphics[width=\columnwidth,height=4cm]{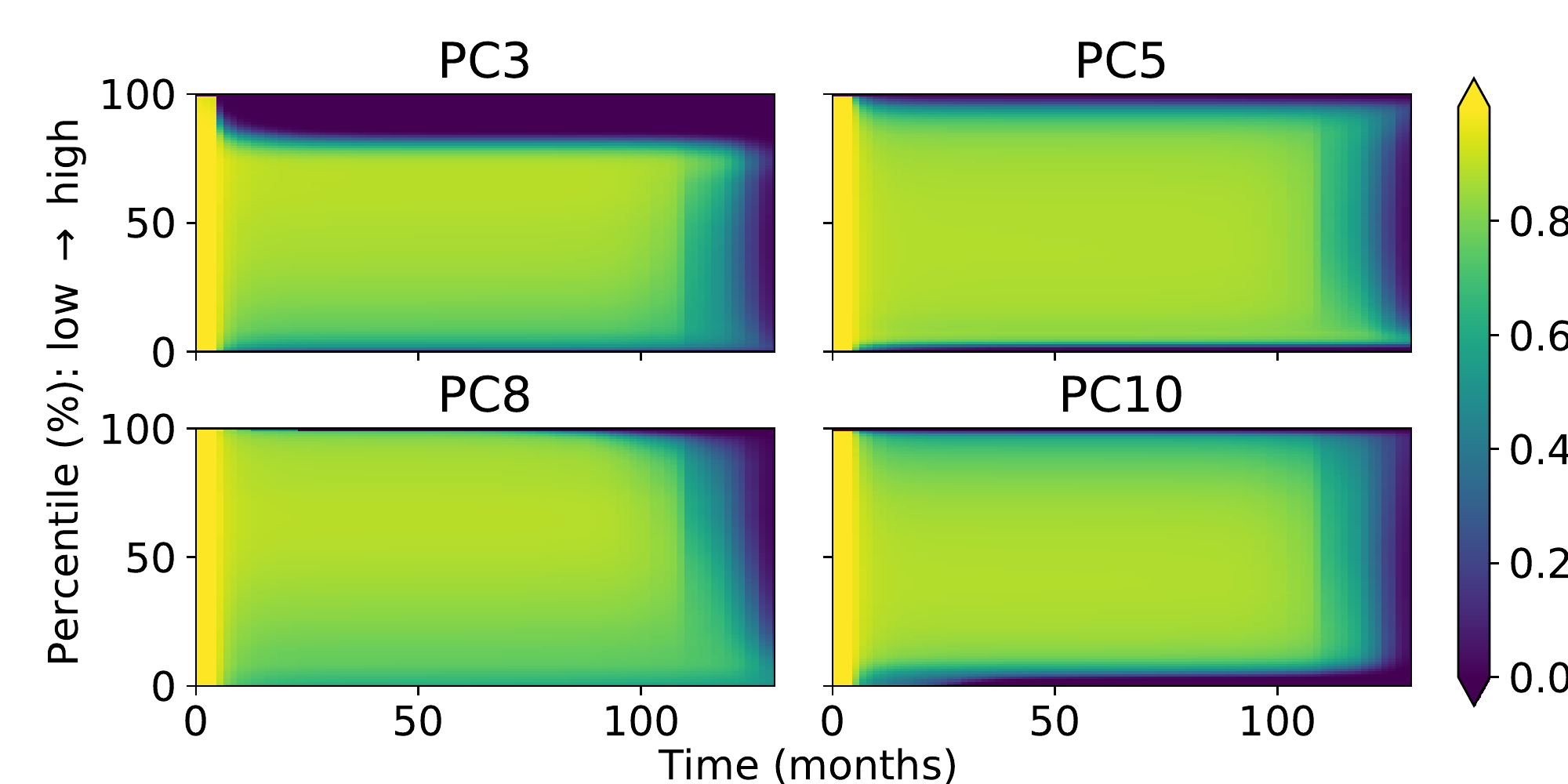}
    \caption{{\bf [Predicting survival]} The contribution of the predictors to the survival probability $S(t|{\bf x})$ with remaining features fixed.}
    \label{fig:density_sur_app}
\end{figure}

The additional results are shown in Figure~\ref{fig:density_sur_app}.

\end{document}


\appendix
\section{The Reddit Network Corpus}
\label{sec:supplemental}

The detailed information of the Reddit Network Corpus is given in this section. The code and data will soon be released to the public.

\subsection{Neologisms}
Table~\ref{table:exp_neo} shows some samples of most frequent and least frequent neologisms in Reddits. These linguistic innovations were collected from \texttt{NoSlang.com}\footnote{\url{https://www.noslang.com/dictionary/}} and \texttt{Urban Dictionary}\footnote{\url{www.urbandictionary.com}}. As explained in the main text, we filtered out lexical entries that: 1) span more than one word, 2) can be found as an entry in an English dictionary after lemmatization, 3) are identified as person names, 4) contain non-alphabetical characters, numbers or emojis and 5) do not show up in our Reddit dataset.

We set loose criteria for word inclusion. Many of the frequent neologisms have already been incorporated into daily lexicon, such as {\it wiki}, {\it google} and {\it instagram}. We manually filtered out these words in our wordlist and the number of such words are less than 100.   
We also keep typos in the curated list, as these words often carry special meanings. For example, {\it alot}, {\it atleast} and {\it recieve} are the typos that are used more than 1 million times, so frequent that they carry some special meanings and functions such as identity assertion. 

As a validation check, we tested some of the results on full list of 170k words, which were created by setting the frequency threshold to 2. We obtained qualitatively similar results.

In terms of ethical concerns, a great number of low frequency words are offensive. These words do not reflect our values. Nor do we endorse the use of these words. However, in order to recover the lexical landscape in Reddit as realistic as possible, we collect the word usage data as they are.

\begin{table}[h]
\centering
\begin{tabular}{ll}
\hline Frequency  & Neologisms \\ \hline
\shortstack{Most frequent \\\\\\\\\\\\\\\\\\\\\\\\}
            & \shortstack[l]{{\it lol, /r, kinda,bitcoin} \\
{\it idk,lmao, tbh,tl;dr}\\
{\it alot, /s, omg, lvl}\\
{\it hahaha, dota, iirc, meh}} \\ \hline
\shortstack{Least frequent  \\\\\\\\\\\\\\\\\\\\\\\\} & \shortstack[l]{{\it thugmonster, blein, sotk,} \\
                                {\it f'tang,boydem,yobbish，}\\
                                {\it ferranti,sonse,yampy,} \\
                                {\it newbiggin}}     \\

\hline
\end{tabular}
\caption{Examples of neologisms.}
\label{table:exp_neo}
\end{table}

\subsection{Inter-community graph}
In order to validate our approach to construct the inter-community graph, we constructed different inter-community graphs by setting the threshold to 2, 3 and 4. One concern is that setting the threshold too low (>=1) results in extremely dense graphs, which are challenging to process. 
After extracting the network features from these networks, we compared them by computing the Kendall rank correlation coefficients between these features. The results show that these networks are highly correlated in most features. So adjusting the threshold does not significantly bias our results qualitatively.

\subsection{Network statistics}
Table~\ref{table:general_stats} provides some general statistics of the whole Reddit Network Corpus. 

\begin{table}[tbh]
\centering
\begin{tabular}{lr}
\hline  & Total \\ \hline
Months  & 152  \\
Subreddits & 4420  \\
Inter.Networks & 152  \\
Intra.Networks & 289170   \\
Users & $>$ 30 millions\\
Neologisms & 172k  \\
\hline
\end{tabular}
\caption{Summary of the whole corpus.}
\label{table:general_stats}
\end{table}

\begin{table*}[tbh]
\centering
\begin{tabular}{lrr}
\hline  &  Median & Inter-quartile range\\ \hline
Nodes & 765 & [351,  1776]\\
Density & 0.0059  & [0.0025,  0.0121] \\
Average Degree & 4.19 & [3.04,  6.29] \\
Largest Connected Component & 88.6\% & [81.9\%, 92.4\%] \\
Singletons & 9.4\% & [6.5\%, 13.9\%] \\
Inter-Community Degree & 376 & [146,  1160]   \\
\hline
\end{tabular}
\caption{Statistical Summary of 289170 Subreddit networks.}
\label{table:graph_stats}
\end{table*}

The average duration of the 4420 subreddit communities is 65 months. Statistical summaries of all 289170 networks are presented in Table~\ref{table:graph_stats}.

\section{Correlations between variables}
We took measures to handle confounding variables. Originally we had tested models with or without log transform. Without log transform, the correlations between variables were weak but the all model performance dropped. So finally we adopted the log transform of some variables. 
In Section, we tested each factor individually, so there are no confounds between variables. 
In Table 1, we only evaluate model predictions, but not its feature importance, so correlations between variables do not matter. 
In Section 6.4., we also run the same model across different settings and interpret models cautiously (line 631-635). To verify the effect, we also tested multiple models with one variable at a time, acquiring the same direction of results for key variables.

\begin{figure*}
    \centering
    \includegraphics{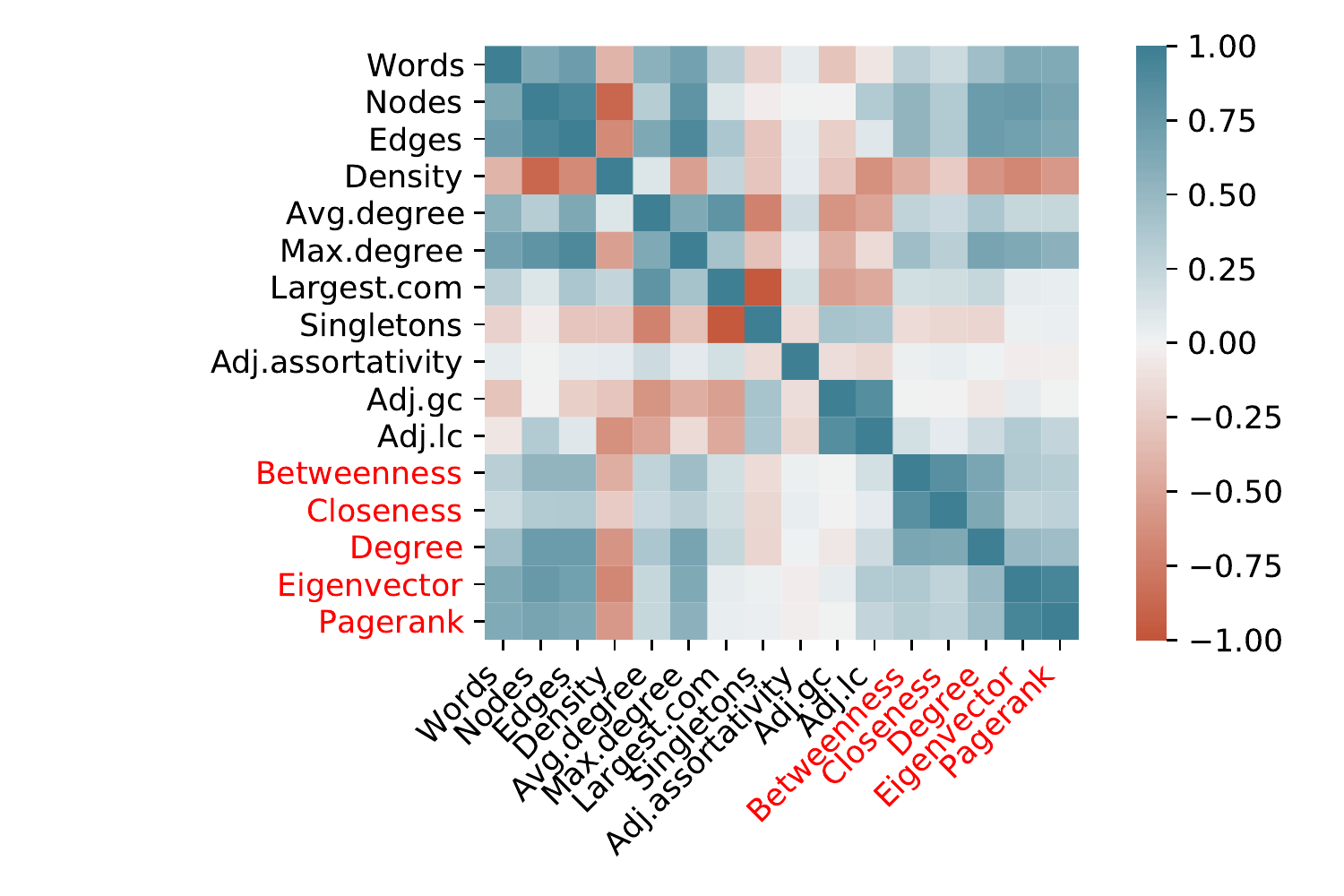}
    \caption{The correlation matrix between variables. Correlation coefficients were computed using the Spearman correlation, as some variables are related in a log-linear relation. Labels in dark are within-community features while labels in red are inter-community features.}
    \label{fig:my_label}
\end{figure*}

\section{Predicting lexical innovations}
In this section, we describe the implementation details for the innovation prediction experiment in Section 5.2. 

\subsection{Feature preprocessing}\label{fpro}
We used mean-variance normalization to normalize all prediction features. Since the distribution of some features are highly skewed, before normalization, we log-transformed the following intra-community features: {\it number of nodes, number of edges, density, average degree, maximum degree}, and the following inter-community features: {\it degree centrality, closeness centrality, Pagerank centrality, betweenness centrality, eigenvalue centrality}. The rest of the features were directly normalized. Whether to perform log-transformation was determined by visual inspection of the density plot. A small number $10^{-6}$ was added before taking the logarithm to improve numerical stability. We found that such a practice improved the performance during cross-validation than directly normalizing all features.

Then these features are partitioned into three different groups to predict the number of innovations per-month.
\begin{itemize}
    \item {\bf Inter.}: {\it degree centrality, closeness centrality, Pagerank centrality, betweenness centrality, eigenvector centrality}
    \item {\bf Intra.}: {\it number of nodes, number of edges, density, average degree, maximum degree, proportion of the largest connected components, proportion of singletons, assortativity, transitivity, average clustering coefficient, adjusted assortativity, adjusted transitivity, adjusted clustering coefficients}.
    \item {\bf All.}: all of the above features.
\end{itemize}
Some of the features were correlated and the correlations varied from weak to strong. Since the goal of this experiment was to explore to what extent innovation number could be predicted with only network features, we did not attempt to de-correlate these features. However, we found that adding more features did increase the performance.  

\subsection{Implementation}
All models were implemented in \texttt{sklearn}. The baseline was the mean number of innovations across all time and all subreddits as the prediction. For the rest of two models, we performed ten-fold cross-validation to select the best parameters. After parameter selection, the regularization parameter for the Poisson regression was $10^{-2}$ and the maximum number of iteration was 300. For the histogram based gradient boosting trees, the maximum number of split was set to 256 and the loss was the Poisson loss. Otherwise we kept the default hyperparameters.

The data were partitioned into training and test sets with a ratio of 90\%/10\%. We run each model 20 times with a different random partition each time. The resulting metrics were averaged across the metrics obtained from the test sets over 20 runs.

\section{Deep survival analysis}
\label{sec:appendix}
In this section, we describe the details of deep survival analysis.
\subsection{Model specification}
We adopt the Logistic Hazard model developed in following works \cite{kvamme2019continuous,JMLR:v20:18-424}. The original derivation comes from \citet{kvamme2019continuous}.

In survival analysis, given a set of discrete time steps $T=\{t_1,t_2,\dots,t_n\}$ and the event time $t^*$, the goal is to estimate the probability mass distribution (PMF) of the event time $f(t)$ and the survival function $S(t)$.

\begin{equation}
    \begin{aligned}
    &f(t) = P(t^*=t_i), \\
    &S(t) = P(t^*>t_i) = \sum_{j>i}f(t_j) 
    \end{aligned}
\end{equation}

The model can also be expressed as the hazard function $h(t)$.
\begin{equation}
    \begin{aligned}
    h(t) &= P(t^*=t_i|t^*>t_{i-1}) \\
        &=\frac{f(t_i)}{S(t_{i-1})} \\
        &= \frac{S(t_{i-1})-S(t_i)}{S(t_{i-1})}
    \end{aligned}
\end{equation}

With the above equations, the survival function can be rewritted as follows.
\begin{equation}
    \begin{aligned}
    f(t_i) = h(t_i)S(t_{i-1}) \\
    S(t_i) = [1-h(t_i)]S(t_{i-1})
    \end{aligned}
\end{equation}

It then follows that
\begin{equation}
    S(t_i) = \prod_{k=1}^i[1-h(t_k)]
\end{equation}

For each individual $i$, the likelihood function can be formulated as
\begin{equation}
    L_i = f(t_i)^{d_i}S(t_i)^{1-d_i}
\end{equation}

The above equation can be rewritten with respect to the hazard function.
\begin{equation}
    \begin{aligned}
     L_i =& f(t_i)^{d_i}S(t_i)^{1-d_i} \\
     =&[h(t_i)S(t_{i-1})]^{d_i}\Big([1-h(t_i)]S(t_{i-1})\Big)^{1-d_i} \\
     =& h(t_i)^{d_i}[1-h(t_i)]^{1-d_i}S(t_{i-1}) \\
     =& h(t_i)^{d_i}[1-h(t_i)]^{1-d_i}\prod_{k=1}^{i-1}[1-h(t_k)]
    \end{aligned}
\end{equation}

The loss function is negative log likelihood function, the negative of the sum of  $log(L_i)$ over all samples. After some algebraic operations, the loss function of the Logistic Hazard model can be formulated as the common binary cross-entroy function. 

\begin{equation}\label{eq:loss}
    \begin{aligned}
            \mathcal{L}=-\frac{1}{n}\sum_{i=1}^n\sum_{j=1}^T\big(y_{ij}log[h(t_j|\textbf{x}_i)]\\ +(1-y_{ij})log[1-h(t_j|\textbf{x}_i)]\big)
    \end{aligned}
\end{equation}
where $y_{ij}$ is the binary event indicator for sample $i$ at time $t$.

Let ${\bf x}$ be an input feature vector and $\phi({\bf x})\in \mathbb{R}^h$ is the neural network that transforms input ${\bf x}$ into $h$ output vectors. Each output vector corresponds to a discrete time step such that $\phi({\bf x})=\{\phi_1({\bf x}),\phi_2({\bf x}),\dots,\phi_h({\bf x})\}$. The hazard function then can be approximated by the sigmoid function.

\begin{equation}
    h(t_i|{\bf x}) = \frac{1}{1+exp[-\phi_i({\bf x})]}
\end{equation}

\subsection{Implementation}
Models of deep survival analysis were implemented using the package \texttt{pycox} \cite{JMLR:v20:18-424}\footnote{\url{https://github.com/havakv/pycox}}.

The network features were normalized and partitioned in the same way as described in Section~\ref{fpro}.

The actual survival time for these neologisms varies from 3 to 152 months. First, we discretized the survival time measured in actual months into 100 intervals based on the distribution of the event times, with the assumption that each interval has the same decrease of the survival probability. The resulting grid will be denser during months with more event times and sparser during months with less event times. Such a practice is recommended by \citet{kvamme2019continuous}, as it reduces parameters and stabilizes training.

We trained a three-layered Logistic Hazard model. For each of the first two layers, we used a linear layer with 256 hidden dimensions and ReLU activation function, follow by batch normalization and a dropout with a probability of 0.1. The last layer was a linear layer with output dimension of 100 followed by a sigmoid activation function.

During training, we used the Adam optimizer with a learning rate of 0.001 and a batch size of 2048 samples. All hyperparameters were tuned with a simple grid search on the development set. Each model was trained for 5 epochs and was run 10 times with different random seeds and different partitions of data each time. The performance metrics were averaged over all 10 runs. These models were trained on a Nvidia V100 GPU and each run took about less than a minute to complete. 

In each run, the data were randomly partitioned into around 80\%, 10\% and 10\% portions as training, development and test sets with different random seeds. In order to avoid information leaking, we ensured that samples in these three sets were from distinct subreddits.

\subsection{Baseline models}
We also run baseline Cox’s proportional hazard models with the same data partitions and discretization scheme. The Cox's model estimates the hazard function $h(t_i|{\bf x})$ with the following equations.

\begin{equation}
    h(t_i|{\bf x}) = \beta_0(t_i)exp\Big(\sum_{i=1}^n\beta_i({\bf x}-\overline{{\bf x}}) \Big)
\end{equation}

We run the model ten times and report the average performance. All baseline Cox's models were implemented using the \texttt{CoxPHFitter} function via the package \texttt{lifelines}\footnote{\url{https://github.com/CamDavidsonPilon/lifelines}}.

\subsection{Additional Analysis}
\begin{figure}[tbh]
    \centering
    \includegraphics[width=\columnwidth]{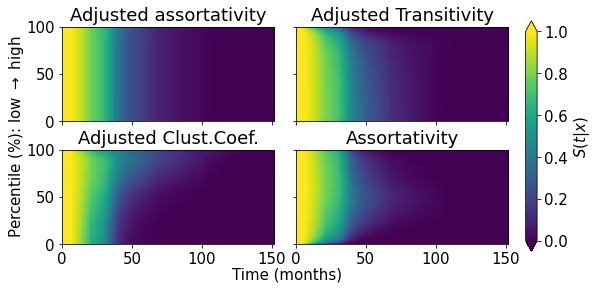}
    \includegraphics[width=\columnwidth]{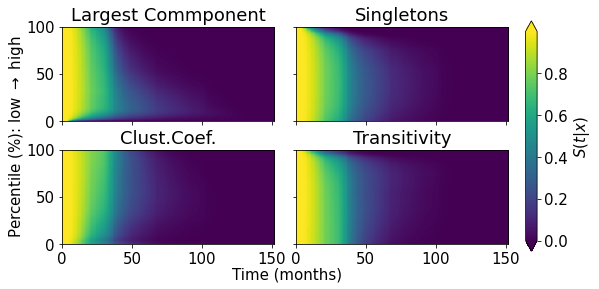}
    \includegraphics[width=\columnwidth]{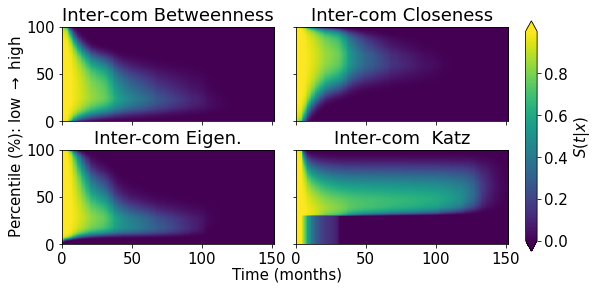}
    \caption{The contributions of individual features to the survival probability $S(t|{\bf x})$ across time steps.}
    \label{fig:variable_all}
\end{figure}

Here we present the contributions of other variables in Figure~\ref{fig:variable_all}. In general, changes of inter-community features alters survival probability more than changes of intra-community features, suggesting that inter-community features are more effective, a result consistent with the evaluation.

Interpreting the contributions of individual features can be challenging as there might be non-linear interactions that complicate a straightforward interpretation. However, we find that these effects are quite consistent across logistic hazard models trained on all features, and also match the directions of coefficients in the Cox's models under the same settings, though some variations still exist.

\bibliography{anthology,custom}
\bibliographystyle{acl_natbib}